\documentclass[10pt,twocolumn,letterpaper]{article}
\usepackage[pagenumbers]{cvpr}
\usepackage{amsmath,amssymb,amsthm}
\usepackage{graphicx,booktabs,xcolor}
\definecolor{cvprblue}{rgb}{0.21,0.49,0.74}
\usepackage[pagebackref,breaklinks,colorlinks,allcolors=cvprblue]{hyperref}
\usepackage{needspace}
\newcommand{\R}{\mathbb{R}}
\newcommand{\Cay}{\operatorname{Cayley}}
\newcommand{\diag}{\operatorname{diag}}
\newtheorem{theorem}{Theorem}
\newtheorem{proposition}{Proposition}
\title{Readout Geometry and Velocity Risk in Rolling-Shutter Rigs under Clock Uncertainty}
\author{%
\begin{tabular}{ccc}
Lunshuo Tian$^{*}$ & Dongheng Liu$^{*}$ & Mingwei Ma\\[-1pt]
\small UNSW Sydney, Australia & \small Sorbonne University, France &
\small South China University of Technology, China\\[-1pt]
\small\ttfamily lunshuo.tian@gmail.com & \small\ttfamily pagniawski@gmail.com &
\small\ttfamily 202520116703@mail.scut.edu.cn\\[-1pt]
\multicolumn{3}{c}{\footnotesize $^{*}$Equal contribution.}
\end{tabular}}
\hypersetup{pdftitle={Readout Geometry and Velocity Risk in Rolling-Shutter Rigs under Clock Uncertainty},pdfauthor={Lunshuo Tian, Dongheng Liu, and Mingwei Ma}}
\makeatletter
\patchcmd{\@maketitle}{\vspace*{24pt}}{\vspace*{4pt}}{}{}
\patchcmd{\@maketitle}{\vspace*{12pt}}{\vspace*{-11.73pt}}{}{}
\makeatother
\begin{document}
\twocolumn[{
\maketitle
\begin{minipage}{\textwidth}
\centering
\includegraphics[width=.94\textwidth]{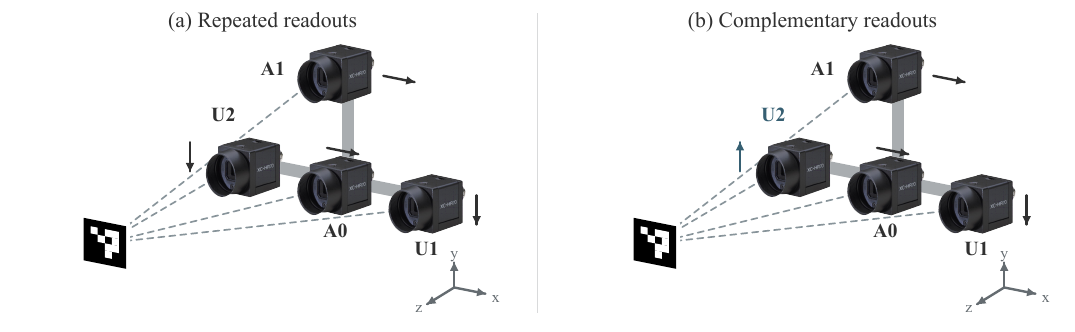}
\captionsetup{hypcap=false}
\captionof{figure}{Oblique view of the nominal rig in normalized body coordinates. A0/A1 form a calibrated pair at $(0,0,0)$ and $(0,.20,0)$. U1/U2 have independently unknown poses and clocks, with nominal centers at $(\pm.12,0,0)$. Their rolls are $(+90^\circ,+90^\circ)$ in (a) and $(+90^\circ,-90^\circ)$ in (b). Cameras face $+z$. Dashed lines show views of a shared ArUco landmark. Solid arrows follow increasing exposure time. The local ambiguity in (a) uses a shared nonzero lateral ordinate $y$ (Sec.~\ref{sec:ambiguity}). Camera photo by Yingz (\href{https://commons.wikimedia.org/wiki/File:VGR_camera.jpg}{public domain}).}
\label{fig:rig}
\end{minipage}\par\vspace{6pt}
}]

\begin{abstract}
We study how rolling-shutter readout changes velocity estimation with unknown auxiliary poses and clocks. Two moving anchors have known relative calibration and share per-point exposure times. On a fixed local class around lateral mounts with an acceleration cone, repeated readout has minimax squared velocity risk at least of order $\sigma\rho_{\mathrm{eff}}$ and complementary readout has risk of order $\sigma^2$, as noise $\sigma\to0$ at fixed admissible clock displacement $\rho_{\mathrm{eff}}>0$. A finite compensation path and a nonlinear inverse establish this separation at equal observation budgets with all nuisances retained. At $.1$\,px, the same 62-parameter estimator gives $.227$\,m/s RMSE for complementary readout versus $.356/.354$ across 256 worlds from a specified mixture. Simultaneous $95\%$ bounds put both mean squared-error reductions above $.0389$\,(m/s)$^2$. Over the mixture support, exact aliases with 12 uniformly spaced rows per camera give equal minimax root risk $.4$\,m/s for all layouts.
\end{abstract}

\section{Introduction}
For indoor robots with calibrated stereo and auxiliary rolling-shutter (RS) cameras, we ask how one scan reversal affects short-window metric velocity estimates.

We make this design choice concrete with two moving anchors and two auxiliaries (Fig.~\ref{fig:rig}). The anchors share the same per-point exposure times and a known metric baseline. Their image observations remain noisy. We compare repeated $+90^\circ$ and complementary $+90^\circ/-90^\circ$ auxiliary readouts on the same local physical class, retaining 96 noisy coordinates and all 62 unknowns, including scene points, auxiliary poses and clocks.

For each fixed finite $D>0$, sufficiently small noise-independent physical radii and a fixed admissible clock step $\rho_{\mathrm{eff}}>0$, we prove $R^*_{++}\gtrsim(L/T)^2\sigma\rho_{\mathrm{eff}}$ and $R^*_{+-}\asymp(L/T)^2\sigma^2$ as $\sigma\to0$. The class satisfies $\|\omega\|\leq D\|(\bar a,B)\|$ on a common safe row branch, relating reference angular velocity to translational and angular acceleration. A shared clock shift can hide axial velocity changes in the repeated layout. Complementary readouts would require incompatible acceleration directions from the same rig, yielding a nonlinear velocity inverse. Appendix C.1 gives matching $\min\{a_*^2\rho^2,\sigma\rho\}$ risk on a known two-parameter path family.

Information laws characterize mounting and readout scales near $(\pm3/25,0,0)$. At the reference scene, both layouts have local quadratic risk at generic safe mounts (Appendix J.2). A calibration bound quantifies the metric-uncertainty floor. At $.1$-pixel noise, 256 independent simulated worlds retain the 96-measurement, 62-unknown contract and give finite-sample confidence bounds on the estimator's mean loss. Clock, angular-rate and acceleration interventions examine timing cost and motion dependence. We match observation counts, unknown counts and covariance.

\section{Related work}
RS stereo recovers structure and motion~\cite{ref1,ref8,ref10}. Albl et al.~\cite{ref2} analyze parallel-readout degeneracy and demonstrate mixed readout directions with handheld phone footage. Albl et al. and Zhong et al.~\cite{ref4,ref5} use synchronized reverse scans for GS reconstruction. HaRP~\cite{ref28} uses four coaxial channels for HDR video. Bapat et al.~\cite{ref27} validate calibrated multi-camera row tracking against external poses. Ito--Okatani~\cite{ref3} characterize critical motion through RS self-calibration.

Early RS methods recover object velocity from known 3D points~\cite{ref31} or fit camera velocity in bundle adjustment~\cite{ref32}. With supplied timestamps, Su et al.~\cite{ref18} recover structure and translation under known rotation, including higher Taylor orders. Hruby--Pollefeys~\cite{ref20} estimate full-DoF motion from point tracks, and Pan et al.~\cite{ref21} use optical flow with translation up to scale. We retain independently unknown auxiliary clocks and poses within a shared trajectory and compare finite metric-velocity risk.

Barath~\cite{ref12} adds affine-correspondence constraints to improve velocity conditioning under velocity--baseline coupling. Order-one RS cameras admit rational projection~\cite{ref16}, and single-view RS SfM uses point projections or line curves under polynomial/Cayley motion~\cite{ref11}. Scanline methods recover relative pose from projected lines~\cite{ref17} or absolute pose on a plane~\cite{ref33}. Known line pencils yield instantaneous absolute pose from concurrent binocular scanlines~\cite{ref23} or a radially distorted scanline~\cite{ref22}. We quantify readout-dependent metric-velocity risk from unknown points with uncertain auxiliary poses and clocks.

\paragraph{Covariance and optimization.}
Oth et al.~\cite{ref9} propagate row noise into residual covariance. NW-RSBA~\cite{ref14} develops covariance-aware fitting and tests changes in velocity, image noise, and readout angle. Our matched NW-RSBA-inspired residual adaptation gives essentially the same velocity accuracy as the implicit solver. Appendices G--H examine motion constraints and residual objectives under a shared input and fitting protocol.

\paragraph{Temporal and intrinsic calibration.}
Li--Mourikis and Yang et al.~\cite{ref6,ref13,ref19} analyze timing compensation and degeneracies in visual--inertial calibration. Nguyen--Lhuillier~\cite{ref26} jointly fit structure, poses, offsets, and line delays from multi-camera video. MVIS~\cite{ref15} and RSSC~\cite{ref7} estimate timing and camera calibration. We supply intrinsics, readout laws, and relative anchor calibration to isolate how readout affects velocity risk with unknown auxiliary poses and offsets.

We compare metric-velocity risk under a single readout change, with the same physical class, noisy coordinates and unknowns in both layouts. Appendix I compares the closest observation models.

\section{Observation model and identical resources}
\paragraph{Maps and units.}
There are $N=12$ static unknown material points $P_i\in\R^3$ with supplied cross-view identities. Normalized time $t\in I\subset\R$ determines
\begin{equation}\label{eq:motion}
\begin{aligned}
C(t)&=tv+\tfrac12t^2a\in\R^3,\\
R(t)&=\Cay([t\omega+t^2B/2]_\times)\in\mathrm{SO}(3),
\end{aligned}
\end{equation}
where $\Cay(K)=(I-K)^{-1}(I+K)$ and $R$ maps world displacements into body coordinates. All twelve coefficients $(v,a,\omega,B)$ are fitted. For physical time unit $T$, the target is $V=v/T$ at the body origin and reference time. When lengths are divided by $L$, write $\bar v=v/L$ and $V=(L/T)\bar v$. At the reference instant, physical translational acceleration is $A_{\mathrm{tr},0}=a/T^2$. With $\dot R R^T=-[\Omega]_\times$ in physical seconds, our Cayley convention gives angular velocity $\Omega_0=-2\omega/T$ and angular acceleration $\Gamma_0=-2B/T^2$. The conventions $C(0)=0$, $R(0)=I$ fix coordinates, not an absolute world pose.

Camera $c$ has body-to-sensor rotation $Q_c$, body centre $b_c$, and normalized time offset $\tau_c$. Define
\begin{equation}\label{eq:projection}
\begin{aligned}
H_{ci}(t)&=Q_c\{R(t)[P_i-C(t)]-b_c\},\\
\pi:\{H\in\R^3:H_3>0\}&\longrightarrow\R^2,\\
\pi(H)&=(H_1/H_3,H_2/H_3).
\end{aligned}
\end{equation}
The first image coordinate is the scan coordinate. Its noiseless value $x_{ci}$ is the selected simple root of
\begin{equation}\label{eq:root}
\begin{aligned}
x_{ci}&=\pi_1(H_{ci}(x_{ci}+\tau_c)),\\
\mu_{ci}&=\pi(H_{ci}(x_{ci}+\tau_c)).
\end{aligned}
\end{equation}
All results use a common positive-depth, simple-row continuation branch. This forward mean uses the latent row; observed-row residuals are separate approximations (Appendix G.7). Stacking roots and ordinates defines $\mu_d:\Theta_d\subset\R^{62}\to\R^{96}$ for design $d$. A material point is shared across views; arbitrary samples on image line curves supply different, weaker correspondence information.

\paragraph{Supplied and estimated quantities.}
The anchors are $(Q,b,\tau)=(I,0,0),(I,\beta e_2,0)$, with known $\beta>0$, intrinsics, and row-time laws. Their world centres $C(t)+R(t)^Tb$ move with the unknown rig. Known baseline supplies metric scale. Auxiliary poses use $Q_s=Q_{\mathrm{nom},s}\exp([\psi_s]_\times)$; their actual $\psi_s,b_s,\tau_s$ are independently unknown. Let
\[
Q_+=\begin{pmatrix}0&-1&0\\1&0&0\\0&0&1\end{pmatrix},\qquad Q_-=Q_+^T.
\]
The designs are $(Q_+,Q_+)$ and $(Q_+,Q_-)$. Both retain 36 point, 12 motion, and 14 auxiliary coordinates, with 96 scalar observations counting shared anchors once. One auxiliary gives 55 unknowns and 72 observations; 48 points and two auxiliaries give 170 and 384.

\paragraph{Statistical experiment.}
For known parameter-independent SPD covariance shape $\Sigma$ and normalized noise $\sigma=\sigma_{\mathrm{px}}/f$, whiten to
\begin{equation}\label{eq:experiment}
\begin{gathered}
Y=F_d(\theta)+\sigma Z,\qquad F_d=\Sigma^{-1/2}\mu_d,\\
Z\sim N(0,I_{96}).
\end{gathered}
\end{equation}
Both layouts use the same $\Sigma$. Define $R_d^*=\inf_{\widehat V}\sup_{\theta\in\mathcal P}\mathbb E_\theta\|\widehat V-V\|^2$. We differentiate the unwhitened mean: $J=D_\theta\mu_d=[J_v,J_n]$, with velocity columns $J_v$ and nuisance columns $J_n$. Thus $D_\theta F_d=\Sigma^{-1/2}J$, and the efficient information is
\begin{equation}\label{eq:information}
w^TI_vw=\sigma^{-2}\inf_q\|\Sigma^{-1/2}(J_vw+J_nq)\|^2.
\end{equation}
Profiling retains all 52 nuisance coordinates with one auxiliary and all 59 with two, including singular nuisance spans. In dimensional length coordinates $I_V=T^2I_v$. Inverse information is a local benchmark, not an estimator-error bound.

\section{Same-resource finite-risk separation}
The reference uses twelve noncoplanar rational points, anchor baseline $\beta=1/5$, and auxiliary centres $b_{01}=(3/25,0,0)$, $b_{02}=(-3/25,0,0)$ (Appendix A). These centres are distinct. Let $\xi\in\R^{48}$ collect points and both auxiliary poses. Normalize lengths by fixed $L$ and set
\begin{equation}\label{eq:class}
\mathcal P=\left\{\begin{aligned}
&\|\xi-\xi_0\|\leq\epsilon,\quad\|\bar v\|\leq\epsilon_v,\\
&\|(\bar a,B)\|\leq A_0,\quad\|\omega\|\leq D\|(\bar a,B)\|,\\
&|\tau_s|\leq\rho_s,\quad s=1,2
\end{aligned}\right\}.
\end{equation}
For each fixed $0<D<\infty$, the radii can be chosen positive, sufficiently small, and independent of noise. Both layouts use identical coordinate bounds and covariance. The estimator knows the class and branch, not realized geometry or motion.

\paragraph{Acceleration transients.}
We study short readouts near a low-angular-rate instant. In physical units, the cone in~\eqref{eq:class} becomes
\[
\|\Omega_0\|\leq DT\sqrt{4\|A_{\mathrm{tr},0}\|^2/L^2+\|\Gamma_0\|^2}.
\]
With zero translational and nonzero angular acceleration, $\|\Omega_0\|/\|\Gamma_0\|\leq DT$ bounds the characteristic angular-velocity-change time, as in rotational starts, stops, or reversals. Translational acceleration can also support nonzero angular velocity. The class has full 62-dimensional interior, while excluding nonzero steady rotation when both acceleration terms vanish. This constraint applies at the reference instant; we evaluate the complete trajectory at every exposure row.

\begin{theorem}[Separated-rig, same-resource risk separation]\label{thm:separation}
On this one fixed physical class, for some fixed $\rho_{\mathrm{eff}}>0$ within both clock caps and all sufficiently small $\sigma>0$,
\begin{equation}\label{eq:separation}
\begin{aligned}
R_{++}^*(\sigma)&\gtrsim(L/T)^2\sigma\rho_{\mathrm{eff}},\\
R_{+-}^*(\sigma)&\asymp(L/T)^2\sigma^2.
\end{aligned}
\end{equation}
The risk ratio diverges. The repeated-layout statement is a lower bound, not a matching characterization of its full-class risk.
\end{theorem}

\subsection{A finite ambiguity to first order}\label{sec:ambiguity}
After fixing $\mathcal P$ around the zero-lateral-ordinate reference, choose $y\ne0$ with $\sqrt2|y|<\epsilon$ and base mounts $b_s=(h_s,y,0)$, $h_s=\pm3/25$. Put $g=(-1,1,0)$, $j=(1,1,0)$ and hold anchor coordinates $Y_i$ fixed, with $x_i=Y_{i1}/Y_{i3}$. For amplitude $\alpha$ and clock step $t$, use
\begin{equation}\label{eq:path}
\begin{aligned}
\bar a&=-2y\alpha e_3,\quad B=\alpha g,\quad\omega=\alpha te_1,\\
\bar v&=2y\alpha te_3,\qquad\tau_s=t,\\
P_i&=C_{\alpha,t}(x_i)+R_{\alpha,t}(x_i)^TY_i,\\
\psi_s&=\alpha(-2te_3-t^2j),\\
b_s&=(h_s,y,0)+\alpha(2h_s\,t\,e_2-h_s\,t^2e_3).
\end{aligned}
\end{equation}
Here $C,R$ are the exact maps~\eqref{eq:motion}. Compare $(\alpha,0)$ with $(\alpha,t)$. Both anchor images are exactly fixed. For repeated $++$, the quadratic pose and centre terms cancel the entire first derivative in $\alpha$, including the clock/angular-motion cross term. Smooth implicit rows give, on one fixed safe rectangle,
\begin{equation}\label{eq:pathgap}
\begin{aligned}
\|\Delta F_{++}\|&\leq C\alpha^2|t|,\\
\|\Delta V\|&=2|y|(L/T)\alpha|t|.
\end{aligned}
\end{equation}
Omitting the quadratic terms generally leaves $O(\alpha t^2)$. The cancellation holds pointwise before projection, rather than through subtraction of camera images.

Fix $0<t=\rho_{\mathrm{eff}}\leq\min\{\rho_1,\rho_2,t_{\mathrm{safe}},D\sqrt{2+4y^2}\}$. For small $\alpha$, both endpoints lie in the same physical class, including all point displacements and the angular cone. Choosing $\alpha\asymp\sqrt{\sigma/t}$ bounds Gaussian KL while retaining squared velocity separation of order $(L/T)^2\sigma t$. Appendix C proves the cancellation, finite remainder, and membership bound. Constants are not uniform as $y\to0$ or $D\to0$; the clock caps are fixed in this comparison.

\subsection{Why the complementary design has a finite inverse}
In normalized lengths, the anchor chart is $P_i=C(x_i)+R(x_i)^TY_i$, with $\widetilde b_s=b_s+\tau_s\bar v$. Let $x$ collect its 60 nonclock coordinates and $G(x,\tau)$ be the whitened mean. At the static lateral reference $x_*$, define
\[
\begin{aligned}
A&=[\Sigma^{1/2}D_xG(x_*,0)\ H_1\ H_2],\\
(H_s\eta)_{si}&=-r_{si}D_{si}(\eta_a+2[Y_i]_\times\eta_B).
\end{aligned}
\]
Here $r_{si}=\pi_1(Q_s(Y_i-b_s))$, $D_{si}=D\pi(Q_s(Y_i-b_s))Q_s$, $\eta=(\eta_a,\eta_B)$, and $H_s$ is zero outside auxiliary $s$'s rows; $b_s,Q_s,Y_i$ take their reference values. This $96\times72$ matrix has rank 71. Its one-dimensional kernel has clock-jet blocks
\[
j_1=(-2ye_3,(-1,1,0)),\quad j_2=(-2ye_3,(-1,-1,0)).
\]
Both physical clocks must use one shared acceleration direction $c\in S^5$: their blocks are $(\gamma_1c,\gamma_2c)$. The Gram matrix of $j_1,j_2$ has eigenvalues $2+8y^2$ and $2$, so their joint squared approximation error by a common direction is at least 2. The physical restriction stays uniformly away from this kernel, including at $y=0$, with all 62 physical columns retained.

\begin{theorem}[Finite complementary velocity stability]\label{thm:stability}
On sufficiently small classes~\eqref{eq:class},
\begin{equation}\label{eq:secant}
\begin{aligned}
\|V(\theta)-V(\theta')\|
&\leq C(L/T)\\[-2pt]
&\quad\cdot\|F_{+-}(\theta)-F_{+-}(\theta')\|
\end{aligned}
\end{equation}
for every pair. Uniformly over smaller acceleration ceilings and clock caps, including $A_0=0$, for every $\sigma>0$,
\[
R_{+-}^*\asymp(L/T)^2\min\{\epsilon_v^2,\sigma^2\}.
\]
Constants may depend on fixed $D$, geometry, and covariance.
\end{theorem}

\paragraph{From the kernel angle to arbitrary endpoints.}
Set $\mathsf A=\Sigma^{-1/2}A$ and let $E:\R^{60}\to\R^{72}$ insert nonclock coordinates. The static shear $U_\tau=I+N_\tau$ has nonzero blocks
\[
\begin{aligned}
(N_\tau z)_{\mathrm{jet},s}&=\tau_s(z_{\bar a},z_B),\\
(N_\tau z)_{\psi_s}&=2\tau_s z_\omega+\tau_s^2z_B,\\
(N_\tau z)_{\widetilde b_s}&=\tfrac12\tau_s^2z_{\bar a}\\
&\quad+2[b_s]_\times(\tau_sz_\omega+\tfrac12\tau_s^2z_B).
\end{aligned}
\]
Thus $N_\tau^2=0$, $U_\tau^{-1}=I-N_\tau$, and $D_xG(x_*,\tau)=\mathsf A U_\tau E$. The null's $\bar a,B$ blocks vanish, so transport leaves its jets unchanged; compact clock caps preserve a positive uniform kernel angle.

Write $q=(\bar a,B)$. Split any endpoint difference into a nonclock leg at $\tau_1$ and a clock leg at $x_0$. The first allows arbitrary acceleration-direction changes; the second fixes $q_0$. If $\alpha=\|q_0\|>0$, put $c=q_0/\alpha$ and $\gamma=\alpha(\tau_1-\tau_0)$. At intermediate clock $\zeta$, the reference divided-clock column is $V_s=\alpha^{-1}\partial_{\zeta_s}(U_\zeta Ex_0)$. It has jet $s$ equal to $c$ and zero motion inputs, hence $U_\tau^{-1}V_s=V_s$ for independent $\tau,\zeta$. Averaging gives $\bar W\in\R^{72\times2}$ with zero velocity block and common jet direction $c$. Then
\[
\begin{aligned}
\Delta G&=\mathsf A U_{\tau_1}z+e,\qquad
 z=E\Delta x+\bar W\gamma,\\
z_{\mathrm{jet},s}&=\gamma_sc,\qquad z_v=\Delta\bar v,\\
\|e\|&\leq\varepsilon\|(\Delta x,\gamma)\|
 \leq\varepsilon(1+S)\|z\|,
\end{aligned}
\]
where $S$ bounds the full nonclock correction in $\bar W$. The uniform angle gives $\|\mathsf A U_{\tau_1}z\|\geq\nu\|z\|$. On a sufficiently small safe ambient product, $\varepsilon(1+S)<\nu$, giving $\|\Delta G\|\geq[\nu-\varepsilon(1+S)]\|\Delta\bar v\|$. If $\alpha=0$, the cone forces $\omega_0=0$ and exact clock invariance removes the clock leg. Appendix K.6 proves the identities and uniform remainder; K.1--K.2 verify ordinary-world inclusion. Appendix B converts the inverse into minimax bounds using class-global least squares and a velocity testing pair.

\paragraph{Quantified physical domains.}
We re-centre and certify two separate lateral products at $y=.05$\,m. The far fixture uses Appendix A; the near fixture halves its world $X,Y$ coordinates and quarters its depths, giving $.5$--$1.25$\,m. With $L=1$\,m, $T=50$\,ms and $\Sigma=I_{96}$, both full 62-parameter products retain independent $\pm6$\,ms clocks. Table~\ref{tab:finite-domains} gives their geometric radii and target-secant constants. The respective rotation-vector and speed bounds are $0.9\,\mu$rad/$2.4\,\mu$m/s and $30\,\mu$rad/$80\,\mu$m/s. Appendix K specifies all motion caps, ordinary-world inclusion and finite Gaussian bounds.

For Near, retaining common quadratic coefficients gives contraction $.506687$ and $C<269.363$. At $.1$\,px each layout's minimax risk is close to that of zero prediction on these products (Appendix X). Table~\ref{tab:finite-domains} gives the sufficient low-noise thresholds for certified ordering. Section~\ref{sec:practical} evaluates mean loss on a separate population.

\begin{table}[t]
\centering
\caption{Explicit physical products. $\delta P,\delta b$ are world-point and mounting component radii in $\mu$m. The normalized target bound is $\|\Delta\bar v\|\leq C\|\Delta\mu\|$. Below $\sigma_{\rm ord}$, $+-$ has lower minimax risk than $++$. At $.1$\,px the last column lower-bounds each layout's minimax squared risk as a fraction of the zero-predictor bound.}
\label{tab:finite-domains}
\small
\setlength{\tabcolsep}{2.6pt}
\begin{tabular}{lrrrrr}
\toprule
Fixture & $\delta P$ & $\delta b$ & $C$ & $\sigma_{\rm ord}$ [px] & Risk fraction\\
\midrule
Far & $.6$ & $.3$ & $20000$ & $1.12\!\times\!10^{-13}$ & $.9999999998$\\
Near & $20$ & $10$ & $270$ & $2.95\!\times\!10^{-10}$ & $.99989$\\
\bottomrule
\end{tabular}
\end{table}

\section{Geometric scales and their limits}
The risk comparison has model-specific local information counterparts. For one quarter-turn auxiliary near centre $(b_x,0,b_z)$, evaluate $v=\omega=B=\tau=0$, $a=\alpha(-1+\eta_1,1,\eta_3)^T$, $\alpha>0$. All 55 physical coordinates remain fitted. Put
\[
r^2=(\alpha/L)^2+\eta_1^2+\eta_3^2+(b_x/L)^2+(b_z/L)^2.
\]
\Needspace{5\baselineskip}
\begin{theorem}[Joint motion--mounting scale]\label{thm:mounting}
For the certified world and sufficiently small $r$,
\begin{equation}\label{eq:mounting}
I_v\asymp\sigma^{-2}\diag(r^2,1,1).
\end{equation}
The comparison is uniform over the five perturbation ratios, in the original velocity coordinates, after profiling all 52 nuisances.
\end{theorem}

\begin{figure}[t]
\centering
\includegraphics[width=\linewidth]{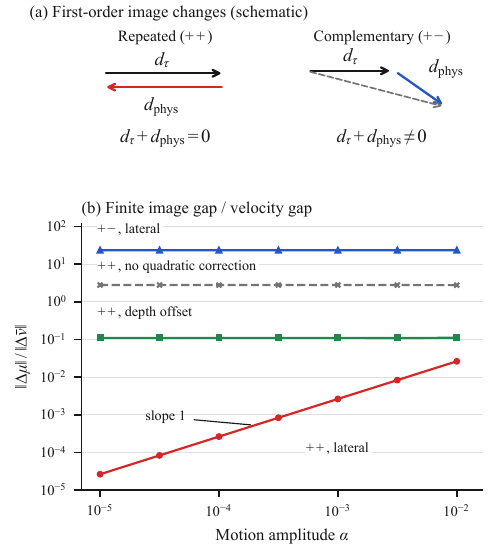}
\caption{Finite compensation and readout sensitivity. (a) At fixed $t$, $d_\tau$ and $d_{\rm phys}$ are the coefficients of $\alpha$ from the clock step and coupled physical updates. They cancel for repeated readouts, while velocity changes to first order. Arrows are schematic. (b) Direct 80-digit evaluation at 13 amplitudes, with lateral ordinate $y=.01$, mounts $(\pm.12,y,z)$, $t=.008$ and $\Sigma=I_{96}$. Depth offset or removing quadratic corrections breaks the cancellation.}
\label{fig:pairs}
\end{figure}

The proof uses a quotient-rank and uniform-remainder argument (Appendix D). For $A=\|a\|>0$, minimizing over acceleration direction gives $\sigma^{-2}[(A/L)^2+(h_\perp/L)^2]$, where $h_\perp=\sqrt{b_x^2+b_z^2}$. Equal altitudes can have very different leading constants (Appendix E). At actual zero acceleration the physical clock column vanishes, so the divided-clock limit describes a different experiment.

For regular roll $\theta$, the exceptional acceleration line is $\operatorname{span}(-\cos(\theta/2),\sin(\theta/2),0)$. For $\theta_\pm=\pi/2\pm\delta/2$, the nearby two-auxiliary law is
\[
\inf_{\|c\|=1}\lambda_{\min} I_{v,\mathrm{full}}(A,c,\delta)
\asymp\sigma^{-2}(A^2+\delta^2),\quad A>0,
\]
with fixed length normalization and all 59 nuisances profiled, including both clocks, uniformly over the ratio of small $A,\delta$ (Appendix F). The crossover is $|\delta|\asymp A$. These are local information statements, distinct from achieved-estimator risk laws.

\paragraph{Reading the mounting transition.}
For the four-camera angular path, lateral spacing preserves the cancellation. A common normalized depth offset $z$ changes the repeated-layout lower envelope, before amplitude saturation, to
\[
R_{\mathrm{rep}}^*\gtrsim(L/T)^2y^2
\min\{\sigma\rho_{\mathrm{eff}},\sigma^2/z^2\}.
\]
For depth offset $z_{\rm m}$ the lower-bound crossover is $\sigma_{\rm px}\asymp f\rho_{\mathrm{eff}}(z_{\rm m}/L)^2$. Appendix V extends this envelope to one imperfect clock report. At the reference scene, an open dense subset of safe mounts gives both layouts $\Theta(\sigma^2)$ on their own small classes (Appendix J.2). Any class containing the zero-depth path retains the repeated lower bound. Known anchor skew lifts the static defect at the rational fixtures (Appendix N). Reprofiled $++$ pairs can hide a $.1$\,m/s gap at $.1$\,px (Appendix O).

\section{Metric calibration and velocity risk}
The main theorem conditions on known anchor calibration. To quantify sensitivity to calibration error, let $F_0$ denote the nominal complementary map on $\mathcal P$, and let the actual mean be $F_0(\theta)+e(\theta)$ with $\sup\|e\|\leq\eta_{\mathrm{cal}}$ in whitened image coordinates.

\begin{proposition}[Calibration perturbation transfer]\label{prop:calibration}
Nominal class-global least squares satisfies
\begin{equation}\label{eq:calibration}
\begin{aligned}
\sup_{\theta\in\mathcal P}\mathbb E\|\widehat V-V\|^2
\leq (L/T)^2\min\bigl\{4\epsilon_v^2,\\
4C^2(96\sigma^2+\eta_{\mathrm{cal}}^2)\bigr\},
\end{aligned}
\end{equation}
where $C$ is the target-secant constant. On a compact safe calibration product, a bounded derivative gives $\eta_{\mathrm{cal}}\leq K\|\kappa-\kappa_0\|$ along an admissible segment.
\end{proposition}
Least squares bounds the fitted mean difference by $2\|\sigma Z+e\|$.
The target secant and expectation give the bound. In tracked-motion renders
at $.1$\,px noise, $+-$ retains the lowest mean loss at A1 transverse
principal-point biases of $\pm.25$ and $\pm1$\,px (Appendix R).

The classical scale ambiguity supplies a necessary limitation. If the anchor baseline is $\beta_0(1+s)$ with unobserved $|s|\leq\delta$, both designs acquire one additional unknown (63 total). For fixed positive class radii and sufficiently small $\sigma,\delta$,
\begin{equation}\label{eq:baseline}
R_{+-}^*(\sigma,\delta)\asymp(L/T)^2(\sigma^2+\delta^2).
\end{equation}
Appendix I.1 proves the scale lower bound. Thus $\delta=O(\sigma)$ preserves the complementary rate, and $\sigma^2+\delta^2=o(\sigma\rho_{\mathrm{eff}})$ guarantees a diverging layout-risk ratio. For lateral $++$, a report of either auxiliary clock or their mean restores quadratic risk with error $O(\sigma)$. Reporting their difference alone preserves the repeated-readout lower bound even at zero reporting error (Appendix U).

\section{Geometric validation and reconstruction}
We test the local rate mechanism on selected pairs and estimate
mean loss at $.1$\,px on independent scenes. Appendix X gives
the worst-case ambiguities on the simulation support.

\paragraph{Connection to the theory.}
Appendix C.2 extends the theorem to $n\geq12$ points with a certified twelve-point subset and safe additional points. The 48-point comparisons fit 170 unknowns.

\subsection{Checks tied directly to the theory}
Exact identities and modular minors give rank 59 for the single-auxiliary mounting certificate and ranks 71 (lateral) and 72 (depth offset) for both two-auxiliary layouts (Appendices A, J.2). We evaluate finite physical pairs and controls without quadratic terms to check the predicted image-gap orders (Fig.~\ref{fig:pairs}).

\begin{figure}[t]
\centering
\includegraphics[width=\columnwidth]{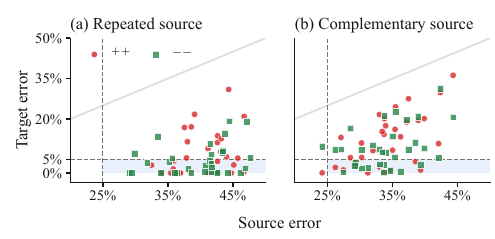}
\caption{Equal-prior error at $.1$\,px. We select pairs on
repeated layouts (a) or complementary (b) and fix both states
for transfer. Shading marks source error $\geq.25$ and target
error $\leq.05$. Counts in $++/--$ order are $18/32,20/32$
in (a) and $10/32,11/32$ in (b) (Appendix L.10).}
\label{fig:known-pairs}
\end{figure}

\paragraph{Finite ambiguity at $.1$\,px.}
The corrected-path pairs have a $.05$\,m/s gap, a $4$\,ms
clock displacement and $y=.2$\,m. Their equal-prior errors
are $.406/.441$ on their respective repeated layouts and below
$.001$ on $+-$. Repeated root-risk lower bounds exceed
$.0225$\,m/s (Appendix L.5).
In 32 reused worlds, we select each layout's closest pair
among nine source fits with gap at least $.05$\,m/s.
Reprojecting both states between repeated and complementary
layouts lowers error for every pair (Fig.~\ref{fig:known-pairs}).
For all 64 repeated-source pairs, the opposite repeat gives
lower error than the complementary target. Of 25 worlds with a
qualifying repeated-to-complementary pair, 15 still attain
the $1.6$\,m/s complementary velocity diameter (Appendix W).

Pooling 27 fits per world preserves the hardest pairs and
yields jointly compatible pairs in $2/32$ worlds (Appendix L.10).
Axial steps of $\pm.025$\,m/s about simulated truth give pairs
fitting all three observations in $28/32$ worlds (Appendix W).

\subsection{Checks inside a certified domain}\label{sec:finite-noise}
Appendix K fits all 62 coordinates within a fixed Far subset at $10^{-11}$--$10^{-8}$\,px, above Far's certified ordering ceiling. The lateral repeated confidence set contains both constructed endpoints in 90/96 noisy four-view draws, versus 0/96 for complementary and depth-offset repeated readouts. This tests the predicted ambiguity on selected pairs.

\begin{figure*}[t]
\centering
\includegraphics[width=.94\textwidth]{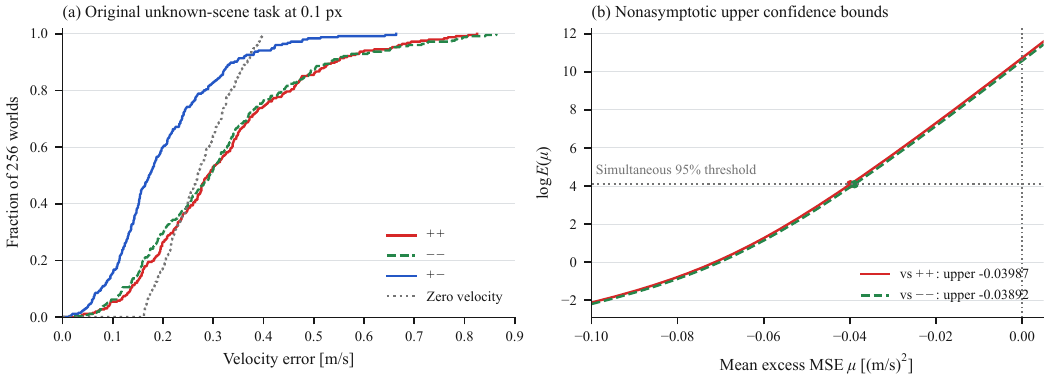}
\caption{Original-task confirmation on 256 new twelve-point worlds at $.1$\,px, with all 62 unknowns and the original two-start budget. (a) Empirical velocity errors, including a descriptive zero predictor. (b) Fixed-mixture e-values for the two mean excess losses. Their threshold crossings give upper confidence endpoints $-.03987/-.03892$\,(m/s)$^2$. Together with the complementary risk upper bound, coverage is at least $95\%$ simultaneously, without bootstrap or asymptotics (Appendix L.12).}
\label{fig:finite-risk-confirmation}
\end{figure*}

\begin{figure*}[t]
\centering
\includegraphics[width=.96\textwidth]{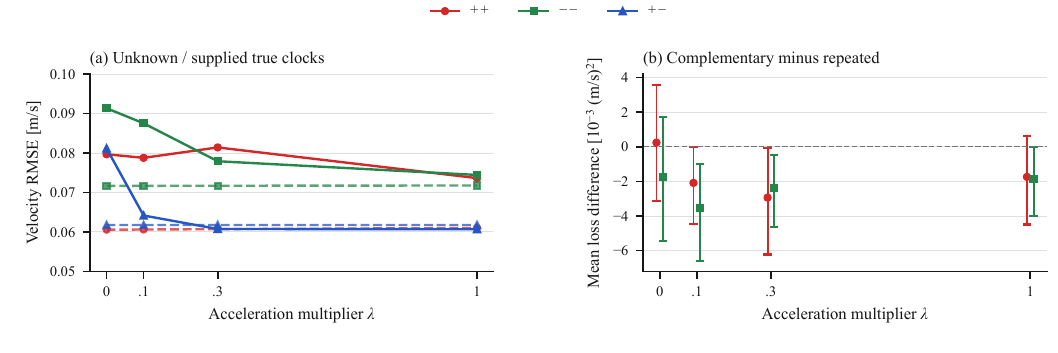}
\caption{Acceleration--timing scan on 32 new worlds at $.1$-pixel noise and $.5$\,rad/s reference angular speed. We scale both acceleration coefficients by $\lambda$, preserving each world's other parameters and noise. (a) Solid/dashed curves fit unknown/supplied true clocks. Lines join sampled levels. (b) Unknown-clock complementary-minus-repeated loss differences with paired 97.5\% world-bootstrap intervals. The two primary contrasts use $\lambda=0$. This scan includes motions outside the theorem cone. Other levels and RMSE curves are descriptive.}
\label{fig:practical}
\end{figure*}

\subsection{Independent-scene finite reconstruction}\label{sec:practical}
\paragraph{Matched simulation protocol.}
The 256-world confirmation uses 128 draws from each of
two motion families. The first perturbs accelerations around
the repeated-layout compensation path. The second samples
acceleration directions on a sphere. Both enforce the same
truth-only physical validity checks for all layouts (Appendix L.12).
Points lie at 2--5\,m and true clocks within $\pm4$\,ms.
Each world supplies the same truth and Gaussian noise draw
across layouts. We initialize structure from noisy anchors, fit
all 62 unknowns with two truth-independent starts and 300
evaluations per start, and select by image residual.

\paragraph{Mean reconstruction loss at $.1$\,px.}
We evaluate the fixed estimator on 128 worlds per family,
conditioned on joint physical validity within each family.
All 768 selected fits stop before the budget and meet the
$10^{-5}$ projected-gradient threshold. Prespecified e-value
inversion~\cite{ref24} gives simultaneous $95\%$ upper bounds
of $-.03987/-.03892$\,(m/s)$^2$ on the mean excess losses,
and $.08068$\,(m/s)$^2$ on complementary expected loss
(Fig.~\ref{fig:finite-risk-confirmation}). Against zero,
equal-mixture upper endpoints are $+.00649$ raw and
$-.01874$\,(m/s)$^2$ after speed projection (Appendix L.12).

\paragraph{Mixture sensitivity.}
For the same two simulated laws at $.1$\,px, post-hoc $95\%$
bounds favor the raw $+-$ fit over both repeats for near-hard
shares above $.123$. With common $.4$\,m/s speed projection,
a separate $95\%$ analysis favors $+-$ over both projected
repeats and zero for every share. A separate paired analysis
gives win-probability lower bounds above $.599$ against either
repeat or zero across Table~\ref{tab:practical}'s six rows
(Appendix L.12).

\Needspace{4\baselineskip}
\paragraph{Velocity ambiguity with separated rows.}\label{sec:practical-geometry}
If point $i$ exposes only at $r_i,s_i$ with $r_i+s_i=k\ne0$,
the changes $(\Delta P_i,\Delta v,\Delta a)=(r_is_iw,kw,-2w)$
cancel at both times. Appendix X constructs identical images at
$\pm.4$\,m/s with known clocks, spread above $.24$\,m and
12 rows per camera spaced by $16.896$\,px over $12.1$\,ms.
Minimax root risk equals $.4$\,m/s on the mixture support
and is at least $.21$\,m/s on the random-direction support.
On the broader class allowing anchor skew, this risk exceeds
$.355$\,m/s at $.1$\,px isotropic noise for every layout and
supplied clock tuple with A1 skew up to $1$\,ms and auxiliary
shifts up to $.25$\,ms each (Appendix O). Appendix P gives
a rank test and an information bound.

\paragraph{Scene contraction.}
In 64 matched new scenes, contraction to $.05$ of the original spread reduces median anchor-derived spread from $.347$ to $.017$\,m. Complementary RMSE rises from $.220$ to $.841$\,m/s. Zero velocity outperforms every layout at that scale, where $162/192$ fits reach the budget cap (Appendix T).

\paragraph{Matched-resource gains.}
In a separate 32-world twelve-point cohort with the same fit budget, complementary RMSE is $.21250$ versus $.34816/.37634$\,m/s (Appendix L.7).

With 48 points, complementary RMSE is 25.2--32.4\% lower than repeated RMSE. It is $.0592$\,m/s versus isotropic zero at $.2892$, and $.0597$ versus the shifted cohort's mean constant at $.0415$ (Appendix L.2).

Six extra starts preserve both paired gains (Appendix L.9).

\begin{table}[t]
\centering\small
\caption{Fixed-estimator velocity RMSE [m/s] at $.1$\,px for two independent cohorts. Rows share worlds within each cohort. Asterisks mark prespecified test conditions. Projection uses the common $.4$\,m/s speed cap. Supplied clocks remove two fitted coordinates.}
\label{tab:practical}
\setlength{\tabcolsep}{2.2pt}
\begin{tabular}{lllrrrr}
\toprule
Cohort & Clocks & Output & $++$ & $--$ & $+-$ & Zero\\
\midrule
Mixture & Free & Raw$^*$ & .356 & .354 & .227 & .279\\
Mixture & Free & Projected & .296 & .293 & .204 & .279\\
\midrule
Random & Free & Raw & .375 & .374 & .226 & .285\\
Random & Free & Projected & .302 & .303 & .205 & .285\\
Random & True & Raw & .321 & .332 & .202 & .285\\
Random & True & Projected$^*$ & .267 & .279 & .184 & .285\\
\bottomrule
\end{tabular}
\end{table}

\subsection{Timing and motion interventions}\label{sec:clockcontrol}
\paragraph{Supplying true auxiliary clocks.}
Supplying both clocks shrinks the mixture gaps by 31\% and
28\%. Post-hoc projected axial and transverse losses favor $+-$.
The raw reduction is larger transversely, with per-axis attenuation
unresolved. Mean and half-difference reports leave 61 unknowns
and raw $++$ RMSEs of $.326/.346$\,m/s, with inconclusive
projected total-loss contrasts (Appendix U).
On the Random cohort, projected upper bounds against both
repeats and zero are negative in both clock protocols.
Paired gaps shrink by $.01155$ $[.00167,.02165]$ against $++$
and $.00610$ $[-.00299,.01525]$ against $--$ (nominal intervals,
Appendix Y).

\paragraph{Nonzero anchor timing.}
In 128 new twelve-point worlds with A1 skew $.5$ or $1$\,ms,
estimated timing yields $.218$ versus $.356/.353$\,m/s RMSE.
All four mean-loss upper bounds are negative at joint 95\%
confidence across supplied or fitted timing (Appendix P).

\paragraph{Angular-rate and acceleration controls.}
With independent angular-rate bounds of $\pm1.6$\,rad/s, complementary RMSE remains $.05916$\,m/s. A $.5$\,rad/s shift takes 14 truths outside the cone and yields $.05914$ versus $.08755/.08085$\,m/s, with both paired intervals below zero. At zero reference acceleration, both primary intervals cross zero and clock-bound activity rises (Fig.~\ref{fig:practical} and Appendix L).

\subsection{Tracked-motion and image checks}\label{sec:motioncontrols}
\paragraph{Layout choice on tracked motions.}
We render 60 motions outside the $D=.15$ cone from ten TUM
sequences~\cite{ref25} (Appendix Q). At $.1$\,px, $+-$ gives
$.199$\,m/s RMSE versus $.343/.367/.293$ for $++/--/$zero.
With TUM points triangulated from SIFT tracks~\cite{ref30}
on 37 of those windows, reversing one scan reduces RMSE by
$7$--$21\%$ relative to repeated readouts ($.497$ versus
$.533/.628$\,m/s). Zero velocity gives $.302$\,m/s. The median
scene spread along the narrowest axis drops from $.313$ to
$.046$\,m (Appendix R).

\paragraph{Scan-slope control on public images.}
On 101 fixed TUM GS/RS image groups~\cite{ref25}, outer
OptiTrack poses supply triangulation scale. Nominal, zero
and reversed slopes give $.80/2.95/3.73$\,m/s equal-sequence
RMSE ($1.01$ for zero velocity). Over 108 attempts, the tracker
secant and held-out RS blend give $.056/.058$\,m/s (Appendix S).

\paragraph{Four-view coverage pilot.}
All 210 coarsely aligned sets per take share $\geq12$ ArUco
corners~\cite{ref29}
(Fig.~\ref{fig:four-view-capture}). D415 stereo timestamps agree.
Scan directions remain unverified. With a $.70$\,ms phone span,
corner RMS falls in B and rises in A. Zero velocity gives the
lowest five-epoch RMSE (Appendices M and Y).

\noindent\begin{minipage}{\columnwidth}
\centering
\includegraphics[width=.92\linewidth]{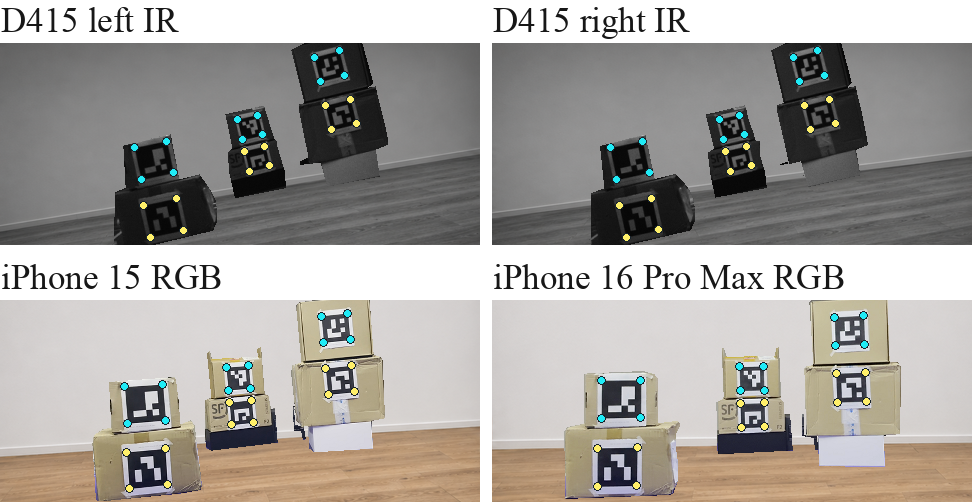}
\captionsetup{hypcap=false}
\captionof{figure}{Four-view coverage in take A. AI-generated backgrounds are for display. Cyan marks velocity-fit corners. Yellow marks D415 stereo reference corners.}
\label{fig:four-view-capture}
\end{minipage}

\section{Conclusion}
On the local motion class, reversing one auxiliary readout breaks
clock--motion compensation and yields minimax squared velocity
risk of order $\sigma^2$ at a fixed observation budget. It lowers
mean loss in paired simulations at $.1$\,px. Replays with
triangulated points preserve the layout ranking, with higher
errors on thinner point clouds. Exact aliases on the wider
support limit recovery. These results inform auxiliary readout
choice for short-window velocity estimation in mobile rigs
with calibrated stereo.

\section*{LLM Usage Statement}
OpenAI GPT drafted text, citations, code, synthetic data scripts, and research claims. GPT assisted with literature search, derivations, experiment design and analysis, ran validation scripts, and provided reviewer-style feedback. GPT and ImageGen generated figure assets. The authors reviewed the outputs and take full responsibility.

\clearpage
\begingroup
\hypersetup{urlcolor=black}

\endgroup


\begin{thebibliography}{33}
\small

\bibitem{ref1} O. Ait-Aider and F. Berry.
\href{https://doi.org/10.1109/ICCV.2009.5459408}{Structure and kinematics triangulation with a rolling shutter stereo rig}.
In \emph{ICCV}, pp. 1835--1840, 2009.

\bibitem{ref2} C. Albl, A. Sugimoto, and T. Pajdla.
Degeneracies in rolling shutter SfM. In \emph{ECCV}, 2016.

\bibitem{ref3} E. Ito and T. Okatani.
Self-calibration-based approach to critical motion sequences of rolling-shutter structure from motion. In \emph{CVPR}, 2017.

\bibitem{ref4} C. Albl et al.
From two rolling shutters to one global shutter. In \emph{CVPR}, 2020.

\bibitem{ref5} Z. Zhong, M. Cao, X. Sun, Z. Wu, Z. Zhou, Y. Zheng, S. Lin, and I. Sato.
Bringing rolling shutter images alive with dual reversed distortion.
In \emph{ECCV}, 2022. arXiv:2203.06451.

\bibitem{ref6} Y. Yang, P. Geneva, K. Eckenhoff, and G. Huang.
\href{https://doi.org/10.1109/LRA.2019.2893803}{Degenerate motion analysis for aided INS with online spatial and temporal sensor calibration}.
\emph{IEEE Robotics and Automation Letters}, 4(2):2070--2077, 2019.

\bibitem{ref7} Y. Zhang, N. Rabbani, B. Liao, C. Wang, Y. Lao, and A. Bartoli.
\href{https://doi.org/10.1007/978-3-032-37550-6_6}{Rolling shutter camera self-calibration}.
In \emph{ECCV}, pp. 96--112, 2026. \href{https://arxiv.org/abs/2608.01509v2}{arXiv:2608.01509v2}.

\bibitem{ref8} O. Saurer, K. Koser, J.-Y. Bouguet, and M. Pollefeys.
Rolling shutter stereo. In \emph{ICCV}, 2013.

\bibitem{ref9} L. Oth, P. Furgale, L. Kneip, and R. Siegwart.
\href{https://doi.org/10.1109/CVPR.2013.179}{Rolling shutter camera calibration}.
In \emph{CVPR}, pp. 1360--1367, 2013.

\bibitem{ref10} B. Fan, Y. Dai, Z. Zhang, and K. Wang.
Differential SfM and image correction for a rolling shutter stereo rig.
\emph{Image and Vision Computing}, 124:104492, 2022.

\bibitem{ref11} S. Err\'{a}zuriz Mu\~noz, K. Kiehn, P. Hruby, and K. Kohn.
Single-view rolling-shutter SfM. arXiv:2603.11888, 2026.

\bibitem{ref12} D. Barath.
\href{https://doi.org/10.1007/978-3-032-37196-6_16}{Rolling shutter relative pose estimation made practical}.
In \emph{ECCV}, 2026. \href{https://arxiv.org/abs/2606.26863v1}{arXiv:2606.26863v1}.

\bibitem{ref13} M. Li and A. I. Mourikis.
\href{https://doi.org/10.1177/0278364913515286}{Online temporal calibration for camera--IMU systems: Theory and algorithms}.
\emph{The International Journal of Robotics Research}, 33(7):947--964, 2014.

\bibitem{ref14} B. Liao, D. Qu, Y. Xue, H. Zhang, and Y. Lao.
Revisiting rolling shutter bundle adjustment: Toward accurate and fast solution.
In \emph{CVPR}, 2023. arXiv:2209.08503v3.

\bibitem{ref15} Y. Yang, P. Geneva, and G. Huang.
\href{https://doi.org/10.1177/02783649241245726}{Multi-visual-inertial system: Analysis, calibration, and estimation}.
\emph{The International Journal of Robotics Research}, 2024. \href{https://arxiv.org/abs/2308.05303v4}{arXiv:2308.05303v4}.

\bibitem{ref16} M. A. Hahn, K. Kohn, O. Marigliano, and T. Pajdla.
Order-one rolling shutter cameras. In \emph{CVPR}, pp.~27007--27016, 2025.

\bibitem{ref17} P. Hruby and M. Pollefeys.
Single-scanline relative pose estimation for rolling shutter cameras.
In \emph{ICCV}, pp.~7143--7153, 2025.

\bibitem{ref18} H. Su, Y. Feng, D. Gehrig, P. Jiang, L. Gao, X. Lagorce, and L. Kneip.
A linear N-point solver for structure and motion from asynchronous tracks.
In \emph{ICCV}, pp. 4339--4348, 2025. arXiv:2507.22733.

\bibitem{ref19} Y. Yang, P. Geneva, X. Zuo, and G. Huang.
\href{https://doi.org/10.1109/TRO.2023.3275878}{Online self-calibration for visual-inertial navigation: Models, analysis, and degeneracy}.
\emph{IEEE Transactions on Robotics}, 2023.

\bibitem{ref20} P. Hruby and M. Pollefeys.
Minimal solvers for full DoF motion estimation from asynchronous tracks.
In \emph{3DV}, 2026. \href{https://arxiv.org/abs/2508.17537v1}{arXiv:2508.17537v1}.

\bibitem{ref21} S. Pan, B. Guan, B. Li, Z. Yu, Z. Liu, Z. Wang, Y. Shang, and Q. Yu.
Minimal solvers for full-DoF motion estimation from asynchronous differential SfM.
arXiv:2606.09218v2, 2026.

\bibitem{ref22} J. C. Dibene and E. Dunn.
Camera resection from known line pencils and a radially distorted scanline.
In \emph{CVPR}, pp. 15843--15851, 2025.

\bibitem{ref23} J. C. Dibene, Y. Maldonado, L. Trujillo, and E. Dunn.
\href{https://doi.org/10.1109/TVCG.2022.3150485}{Prepare for ludicrous speed: Marker-based instantaneous binocular rolling shutter localization}.
\emph{IEEE Transactions on Visualization and Computer Graphics}, 28(5):2201--2211, 2022.

\bibitem{ref24} V. Vovk and R. Wang.
\href{https://doi.org/10.1214/20-AOS2020}{E-values: Calibration, combination and applications}.
\emph{The Annals of Statistics}, 49(3):1736--1754, 2021.

\bibitem{ref25} D. Schubert, N. Demmel, L. von Stumberg, V. Usenko, and D. Cremers.
\href{https://doi.org/10.1109/IROS40897.2019.8968539}{Rolling-shutter modelling for direct visual-inertial odometry}.
In \emph{IROS}, pp.~2462--2469, 2019.

\bibitem{ref26} T.-T. Nguyen and M. Lhuillier.
\href{https://doi.org/10.1016/j.cviu.2017.08.010}{Self-calibration of omnidirectional multi-cameras including synchronization and rolling shutter}.
\emph{Computer Vision and Image Understanding}, 162:166--184, 2017.

\bibitem{ref27} A. Bapat, T. Price, and J.-M. Frahm.
Rolling shutter and radial distortion are features for high frame rate multi-camera tracking.
In \emph{CVPR}, pp.~4824--4833, 2018.

\bibitem{ref28} X. Ji, G. Lin, J. Zhao, Z. Yin, and Y. Zheng.
HaRP: High dynamic range photosequencing through dual reversed shutter scanning.
arXiv:2609.28439, 2026.

\bibitem{ref29} S. Garrido-Jurado, R. Mu\~noz-Salinas, F. J. Madrid-Cuevas, and M. J. Mar\'{i}n-Jim\'{e}nez.
\href{https://doi.org/10.1016/j.patcog.2014.01.005}{Automatic generation and detection of highly reliable fiducial markers under occlusion}.
\emph{Pattern Recognition}, 47(6):2280--2292, 2014.

\bibitem{ref30} D. G. Lowe.
\href{https://doi.org/10.1023/B:VISI.0000029664.99615.94}{Distinctive image features from scale-invariant keypoints}.
\emph{International Journal of Computer Vision}, 60(2):91--110, 2004.

\bibitem{ref31} O. Ait-Aider, N. Andreff, J.-M. Lavest, and P. Martinet.
\href{https://doi.org/10.1007/11744047_5}{Simultaneous object pose and velocity computation using a single view from a rolling shutter camera}.
In \emph{ECCV}, pp.~56--68, 2006.

\bibitem{ref32} O. Grau and J. Pansiot.
\href{https://doi.org/10.1145/2414688.2414700}{Motion and velocity estimation of rolling shutter cameras}.
In \emph{CVMP}, pp.~94--98, 2012.

\bibitem{ref33} F. Bai, A. Sengupta, and A. Bartoli.
\href{https://openaccess.thecvf.com/content/CVPR2022/html/Bai_Scanline_Homographies_for_Rolling-Shutter_Plane_Absolute_Pose_CVPR_2022_paper.html}{Scanline homographies for rolling-shutter plane absolute pose}.
In \emph{CVPR}, pp.~8993--9002, 2022.

\end{thebibliography}
\end{document}


\onecolumn
\begin{center}
{\Large\bfseries Readout Geometry and Velocity Risk in Rolling-Shutter Rigs\\under Clock Uncertainty\par}
\vspace{7pt}
{\large Supplementary Material\par}
\vspace{4pt}
Lunshuo Tian, Dongheng Liu, and Mingwei Ma
\end{center}
\vspace{4pt}
\pdfbookmark[0]{Reading guide}{reading-guide}
\noindent We provide the proofs, final experimental protocols and supporting results for the submitted main paper. The reading guide locates the evidence for each claim; the protocols specify their observation contracts. Exact certificate verifiers and the final Table~2 selected records are in the separate reproducibility archive.

\paragraph{Reading route.}
For the main risk separation, read A--C; for mounting/readout scales,
D--F and J. K gives the finite physical products and certificates;
L.12 gives the $.1$-px, 256-world reconstruction guarantee.
L.5/L.10 and W--Y give finite pairs and broader-support boundaries.
M illustrates the four-view capture; Q--S distinguish tracked-motion
simulation from public-image controls. I and N--P/U--V cover calibration
and supplied timing. Appendix identifiers and reference numbers match
the submitted main paper. The archive README gives code entry points.

\paragraph{Units and evaluation.}
$V=(L/T)\bar v$ and $\sigma=\sigma_{\rm px}/f$.
Errors are RMSE in m/s; squared losses are in (m/s)$^2$.
Experimental inputs and independent units are specified in each protocol.

\vspace{4pt}
\pdfbookmark[0]{Contents}{contents}
{\small\setlength{\parskip}{0pt}
\begin{multicols}{2}
\tableofcontents
\end{multicols}}
\clearpage\twocolumn
\includepdf[pages=1-13,pagecommand={\thispagestyle{plain}},addtotoc={1,section,1,{Derivative construction and exact certificates},app:A,2,section,1,{Finite complementary stability},app:B,4,section,1,{Finite repeated-layout pairs and separation},app:C,7,section,1,{Mounting geometry and clock-critical velocity},app:D,9,section,1,{Leading constants and mounting cost},app:E,11,section,1,{Readout-roll information scales},app:F}]{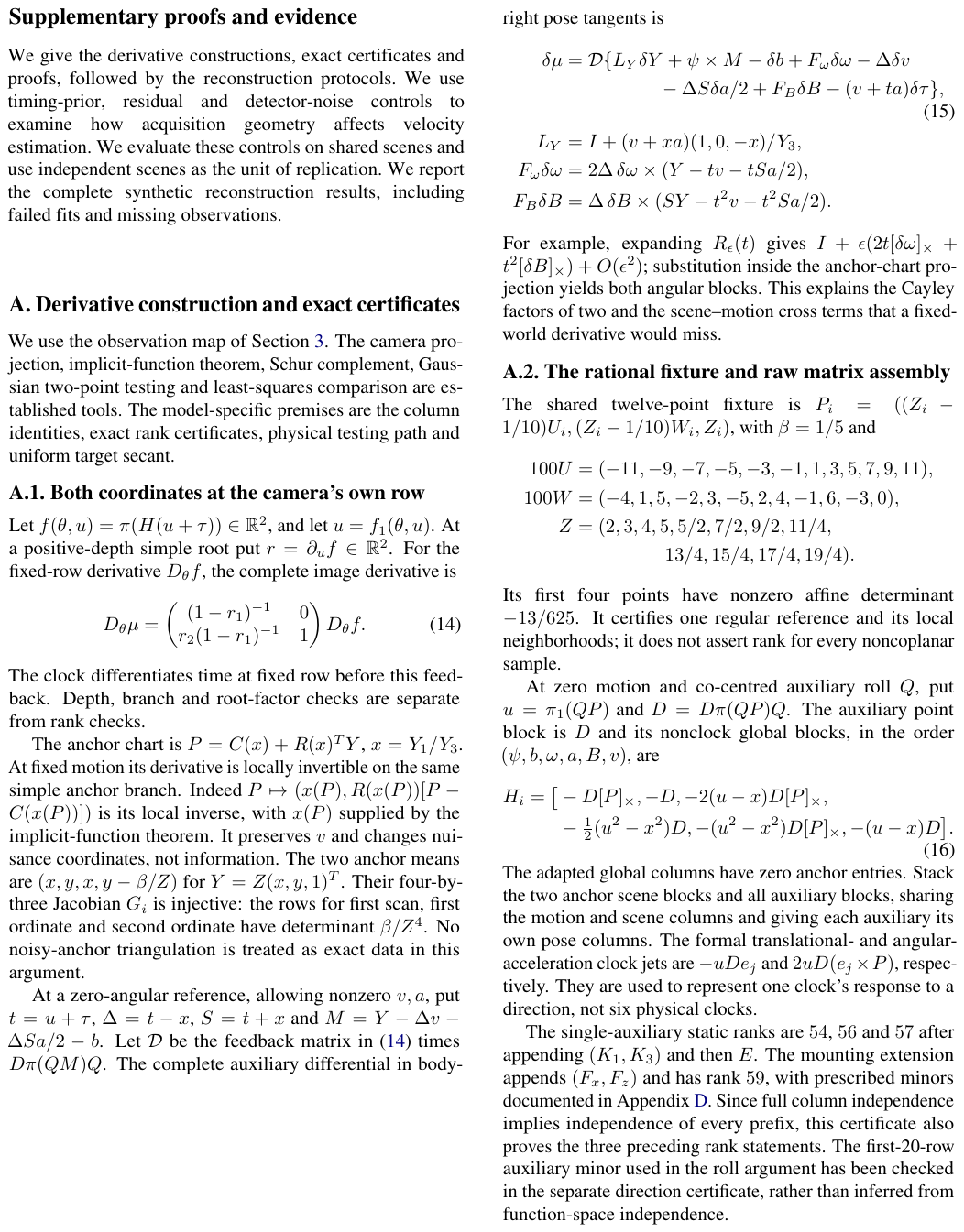}
\setcounter{section}{6}
\section*{F. Readout-roll information scales (continued)}
This residual is $O(|\delta|)$. The actual nonclock columns and
$J_{\tau_s}/A$ differ smoothly from their limits by $O(A)$. The complete
trial residual is therefore $O(A+|\delta|)$, whose squared norm proves
the upper bound. Unbounded clock increments are allowed; the estimate
uses their normalized columns. Common covariance spectral bounds
transfer both raw inequalities before nuisance minimization, including
correlated observations.

The conclusion is a worst-direction scalar information law. At actual
$A=0$ both physical clock columns vanish; nearby static nonclock
subsystems already have positive target information. The punctured
limit in (59) describes a different experiment. Its constants and
neighborhood depend on the reference geometry.

Angle-dependent deformation and perpendicular readouts were studied
by Albl et al.~[2], and their calibrated rig uses known trigger offset
and relative geometry~[4]. Yang et al.~[6] analyze clock--extrinsic--motion
degeneracies. The result here specifies the half-angle critical
acceleration line and complete-nuisance local rate under the stated
anchor contract.

\section{Numerical model and final evaluation protocols}\label{app:G}
\subsection{Implicit-row observations}
We solve each camera's exposure-row equation before evaluating both
coordinates, retaining every motion, structure, pose and clock column.
L specifies the 96/62 and 384/170 measurement/unknown contracts.

\subsection{Exact and finite-scale checks}
A--F/J give rational ranks and high-precision information checks;
K gives domain certificates. L.5/L.10 compare the same complete states
across readouts. The README gives certificate and score-replay commands.

\subsection{Matched camera resources}
Within each comparison, layouts share points, motion, calibration,
correspondences and noise. Extra clock reports are identified explicitly.

\subsection{Rendering for the residual control}
The eight evaluation worlds used in H contain 12 independently placed
textured ArUco panels [29] (IDs 0--11, \texttt{DICT\_5X5\_100}). All 48
corner coordinates are free fitted points; marker dimensions and
coplanarity are withheld. Images have 768 rows and 1024 columns,
focal lengths 768 px and principal point $(383.5,511.5)$ in row--column
order. The anchor baseline is $.2$ m and nominal auxiliary centres are
$(\pm.12,0,.02)$ m. With $T=.05$ s, row $r$ exposes at normalized
time $(r-383.5)/768+\tau$.

We integrate projected texel areas with nearest-surface visibility.
The final residual control uses 1-ms exposure, five temporal midpoint
nodes and independent 2-DN Gaussian intensity noise before clipping
and conversion to 8-bit images. Subpixel corner refinement and marker
IDs supply the observations. This renderer supplies the same detected
pixels to every residual formulation in H.

\subsection{Initialization and optimization}
Fits initialize from noisy anchors and nominal mounts and select by
image residual. H uses QR--LM; L uses bounded TRF. Unsuccessful
terminations remain in scores; KKT tests indicate stationarity.

\subsection{Scores and replication}
RMSE is $\sqrt{n^{-1}\sum_i\|\widehat V_i-V_i\|^2}$.
Each protocol specifies raw/capped/projected loss and its confidence
family. Worlds are independent units; starts and corners are not.

\subsection{Observed-row residuals}\label{sec:observed-row}
For physical parameters $p\in\R^{3n+26}$, normalized observation
$y=(y_1,y_2)$ and $f_p(t)=\pi(H_{c,p}(t))$, define
\[
\begin{aligned}
 e_p(y)&=f_p(y_1+\tau_c(p))-y,\\
 u_p(y)&=\partial_t f_p(y_1+\tau_c(p)),\\
 C_p(y)&=I_2-u_p(y)(1,0),\\
 r_U&=e_p/\sigma,\qquad r_W=C_p^{-1}e_p/\sigma,
 \quad\sigma=.1/768.
\end{aligned}
\]
Covariance standardization follows NW-RSBA~[14], with the present
quadratic-centre/Cayley rigid-rig model. We retain the weight derivative,
\[
 \dot r_W=\sigma^{-1}C_p^{-1}
 (\dot e_p-\dot C_p C_p^{-1}e_p).
\]
At zero residual, the implicit-function identity gives
$D_pr_W=\sigma^{-1}D_p\mu$. A separate $384\times170$ hand Jacobian
checks this identity. Locally, for $y=\mu(p)+n$,
$C_p^{-1}e_p=-n+O(\|n\|^2)$. The Gaussian observation model in the
theory uses latent exposure rows; this observed-row residual is a
finite-noise approximation evaluated separately in H.

\section{Matched NW-RSBA-inspired residual comparison}\label{app:H}
We compare implicit, unweighted observed-row and weighted observed-row
residuals on the same eight rendered worlds in G.4. Every formulation
receives the same detected pixels, correspondences, 170 unknowns,
anchor-derived start and supplied $\pm200\,\mu$s clock bounds.
The shared QR--LM solver allows 200 complete Jacobians, 2000 model
calls and 20 trial steps per Jacobian. Every fit has one start.
All formulations use the same implicit-branch screen; observed-row
residuals also require positive depth and $|1-u_1|>10^{-10}$.

\begin{table}[t]
\centering\small
\caption{Final matched residual control on eight worlds. RMSE is in
m/s. KKT counts are out of eight, in $++/--/+-$ order. All 72 selected
returns enter the scores, including unsuccessful terminations.}
\label{tab:residual-final}
\setlength{\tabcolsep}{3pt}
\begin{tabular}{@{}lrrrr@{}}
\toprule
Residual & $++$ & $--$ & $+-$ & KKT\\
\midrule
Implicit & .06132009 & .05471083 & .03333334 & 8/8/7\\
Observed, unweighted & .06120327 & .05466347 & .03316570 & 6/8/8\\
Observed, weighted & .06132009 & .05471083 & .03333332 & 6/6/6\\
\bottomrule
\end{tabular}
\end{table}

The weighted and implicit formulations give essentially the same
velocity accuracy: the largest paired velocity difference is
$2.6612\times10^{-7}$ m/s. Their complementary reduction against
the better repeated layout is $39.0736\%$. Every method/layout has
14 active bounds among 16 fitted clocks. We recompute each solution
against its own objective because the residual scales differ.

This control isolates readout layout and residual formulation under
our shared trajectory and QR--LM protocol. We adapt the covariance
standardization of NW-RSBA~[14]; its published implementation uses
independent first-order camera motions and Schur acceleration.
The table evaluates this adaptation on matched inputs.

\section{Observation contracts and metric calibration}\label{app:I}
We compare methods through their supplied measurements and unknowns.
NW-RSBA~[14] addresses covariance-aware fitting; MVIS~[15] and rolling
shutter self-calibration~[7] use multi-frame or inertial information.
The RS camera models~[11,12,16] and single-scanline relative pose~[17]
have different correspondence and motion assumptions from the shared
material points, trajectory and unknown auxiliary clocks studied here.

The asynchronous solver of Su et al.~[18] provides linear point
reconstruction and higher-order translation extensions with supplied
angular rates and timestamps. Its homogeneous velocity is up to scale;
supplied acceleration gives a metric variant. Applying it to the present
contract would require those additional inputs and a separated-rig
extension. The main comparison instead fixes the information and
unknowns while changing one auxiliary readout.

\subsection{Proof of the baseline uncertainty corollary}
Fix the main theorem's class $\Theta=\mathcal P_6$ with positive
geometry and velocity radii $\epsilon,\epsilon_v$. Enlarge it to
$\Theta\times[-\delta,\delta]$ by making only the anchor baseline
$\beta_s=\beta_0(1+s)$ unknown. Both designs then have 63 unknowns,
including 60 velocity nuisances. Other anchor calibration and the
fixed covariance remain supplied. Work on a common compact,
positive-depth, simple-row branch for $|s|\leq\delta_{\rm safe}$.

At $\xi_0$, take $\bar v_*=(\epsilon_v/2)e_3$ and
$\bar a=B=\omega=\tau_1=\tau_2=0$. For $\lambda_\pm=1\pm\delta$,
multiply all points, camera centres, baseline and translational velocity
by $\lambda_\pm$, keeping rotations and clocks fixed. At every row,
\[
 H'_{ci}(t)=Q_c\{R(t)[\lambda P_i-\lambda C(t)]-\lambda b_c\}
          =\lambda H_{ci}(t).
\]
Both image coordinates, implicit equations, selected roots and Gaussian
laws therefore agree. The gauge $C(0)=0,R(0)=I$ and units $L,T$ remain fixed.
Put $M^2=\sum_i\|P_i^0\|^2+\sum_{s=1}^2\|b_s^0\|^2$ in the class's
geometry coordinates. For
$\delta\leq\min\{1/2,\epsilon/(2M),\delta_{\rm safe}\}$,
the geometry displacement is at most $\epsilon/2$ and velocity norm at
most $3\epsilon_v/4$. Zero acceleration and angular coefficients obey
the cone. The parallelogram identity under the common law gives
\begin{equation}\tag{60}
 R_d^*(\sigma,\delta)\geq\tfrac14(L/T)^2\epsilon_v^2\delta^2.
\end{equation}
On the safe product let $K_s=\sup\|\partial_sF_{+-,s}\|<\infty$.
Integrating the baseline segment yields
$\|F_{+-,s}-F_{+-,0}\|\leq K_s\delta$.
Nominal global fitting and the velocity inverse give the upper bound
\[
 (L/T)^2\min\{4\epsilon_v^2,4C^2(96\sigma^2+K_s^2\delta^2)\}.
\]
The $s=0$ submodel retains the original $\sigma^2$ lower bound;
together with (60), this proves main Eq.~(13) with fixed-class
constants. The repeated-layout $s=0$ submodel similarly gives
$R_{++}^*\gtrsim(L/T)^2\max\{\sigma\rho_{\rm eff},\delta^2\}$
when its lower pair has positive margins. The scale floor depends on
nonzero admissible velocity; an independently measured metric length
changes the observation contract.

\section{Depth offsets and the boundary of risk separation}\label{app:J}
We give a finite lower envelope for common depth offsets and a
sufficient condition for both layouts to recover regular velocity risk.
All 62 physical unknowns remain.
\subsection{A common depth offset lifts the constructed pair}
Add $ze_3$ to each base mount in (22), retaining the motion, pose and
centre-increment formulas. On a fixed compact safe rectangle in
$(\alpha,t,z)$, set
$G(\alpha,t,z)=F(\alpha,t,z)-F(\alpha,0,z)$.
The exact identities are
\[
 G(0,t,z)=G(\alpha,0,z)=0,\qquad \partial_\alpha G(0,t,0)=0.
\]
Integrating in $z,t$ gives
\[
 \partial_\alpha G(0,t,z)=\int_0^z\int_0^t
 \partial_w\partial_s\partial_\alpha F(0,s,w)\,ds\,dw.
\]
Taylor expansion in $\alpha$, preserving the $t=0$ subtraction in
the remainder, proves
\begin{equation}\tag{61}
 \|G(\alpha,t,z)\|\leq C\alpha|t|\sqrt{\alpha^2+z^2}.
\end{equation}
The constant depends on the fixed points, $h_s,y$, whitening and
branch margins and is uniform on this rectangle. For a class centred
at $(h_s,0,z)$, the C membership calculation is unchanged with safe
bounds uniform in $z$. For a class centred at height zero, replace
the geometry margin by $\epsilon_{\rm phys}-\sqrt{2(y^2+z^2)}>0$.
Fix $t_0\leq\min\{\rho_1,\rho_2,t_{\rm safe},D\sqrt{2+4y^2}\}$
and the corresponding amplitude ceiling $\bar\alpha>0$.
The target gap is $2|y|(L/T)\alpha t_0$.

\paragraph{Proposition 8 (Common-depth lower envelope).}
For every $\sigma>0$,
\begin{equation}\tag{62}
\begin{aligned}
 R_{\rm rep}^*&\geq\tfrac14(L/T)^2y^2t_0^2\min\{\bar\alpha^2,u_*\},\\
 u_*&=\frac{\sqrt{z^4+4\sigma^2/(C^2t_0^2)}-z^2}{2}.
\end{aligned}
\end{equation}
\emph{Proof.} Set $\alpha^2=\min\{\bar\alpha^2,u_*\}$.
Then $\alpha^2(\alpha^2+z^2)\leq\sigma^2/(C^2t_0^2)$,
so (61) bounds Gaussian KL by $1/2$. The squared target gap divided
by 16 gives (62).

Before saturation the envelope is comparable to
$(L/T)^2y^2\min\{\sigma t_0,\sigma^2/z^2\}$, interpreted as
$(L/T)^2y^2\sigma t_0$ at $z=0$. The crossover is
$\sigma\asymp t_0z^2$. Here $z$ is common depth offset, rather than
total triangle altitude. A class containing zero-height hard worlds
retains the slower lower bound even if its nominal height is nonzero.

\subsection{A sufficient criterion for regular mounted rigs}
For safe static mounts form the complete augmented matrix
$A_b\in\R^{96\times72}$ using A.
\paragraph{Theorem 7 (Mounted full-jet sufficiency).}
If $\operatorname{rank}A_b=72$, then for every finite $D\geq0$,
fixed positive physical radii exist around this reference with a
uniform finite velocity secant and
\[
 R_b^*(\sigma)\asymp(L/T)^2\min\{\epsilon_v^2,\sigma^2\}.
\]
\emph{Proof.} For a shared unit acceleration direction $c$, physical
insertion $T_c$ is an isometry, so
$\sigma_{\min}(A_bT_c)\geq\sigma_{\min}(A_b)>0$.
The mounted angular-velocity shear, target-preserving transport and
whitening give the complete-column margin and full nonclock rank in B.
Its centre change, Hadamard division, fixed-matrix local integration,
near/far argument and physical-radius inclusion apply at this reference.
The resulting target secant and Gaussian testing give the risk law.
Raw clocks can remain unidentifiable at zero motion.

At the A fixture with $\beta=1/5$, mounts
$(3/25,0,1/50)$ and $(-3/25,0,1/50)$ give rank 72 in both layouts.
Independent reductions modulo 1009 and 1013 certify the saved
rational matrices without deleting rows or columns. Common smaller
radii give the same physical bounds and regular $\sigma^2$ risk for
both layouts; their constants can differ and are not uniform as
height tends to zero.

Matrix entries are rational in the six mount coordinates. Each
full-rank witness gives a minor with a nonzero polynomial numerator;
the product of the two numerators is nonzero. Simultaneous full rank
is therefore open dense in any nonempty open common safe mounting
domain. This gives regularity on each reference's own small class,
without a common radius across mounts. The pure-lateral family is
exceptional. Full augmented rank is sufficient: the main complementary
result instead uses rank 71 and its physical restriction.

For the lateral mounts $(\pm3/25,y,0)$ at $y=0,1/100$, both layouts
have augmented rank 71. Writing $A(z)$ for common depth shift and
$N$ for its null, $[A(0),A'(0)N]$ has rank 72 at both primes.
Analytic assembly and separate symbolic differentiation agree.
This is a first-order lifting check; the finite lower envelope follows
from the path identities in (61).

\setcounter{section}{10}
\renewcommand{\theequation}{K.\arabic{equation}}
\renewcommand{\thetable}{K.\arabic{table}}
\setcounter{equation}{0}\setcounter{table}{0}

\section{Quantified domains and paired noise}
\label{app:K}

\subsection{Two full-parameter physical products}
We quantify two lateral references with auxiliary centres
$(\pm.12,.05,0)$ and anchor baseline $.2$\,m. The far fixture is the
rational scene in Appendix A. The near fixture replaces each far
reference point $(X,Y,Z)$ by $(X/2,Y/2,Z/4)$; its depths are
$.5$--$1.25$\,m and all four nominal images lie within
$[-.5,.5]^2$. The mounts and calibration are unchanged. These references
are distinct from the $y=0$ centre of the existence theorem.
Throughout this section $L=1$\,m, $T=.05$\,s, $f=768$\,px and
$\Sigma=I_{96}$; normalized noise is $\sigma=\sigma_{\rm px}/768$.

Let $r\in\{F,N\}$ denote the far/near fixture and set
$\nu_F=3\times10^{-5}$, $\nu_N=10^{-3}$.
The fixed ordinary-world product $\mathcal B^r$ is
\begin{equation}\label{eq:box}
\begin{aligned}
 |\bar P_{ij}-\bar P^r_{ij,0}|&\leq .02\nu_r,&|\psi_{sj}|&\leq .03\nu_r,\\
 |\bar b_{sj}-\bar b_{sj,0}|&\leq .01\nu_r,&\|\bar v\|&\leq .004\nu_r,\\
 \|(\bar a,B)\|&\leq .2\nu_r,&\|\omega\|&\leq .15\|(\bar a,B)\|,\\
 |\tau_s|&\leq .12.&&
\end{aligned}
\end{equation}
All point, motion, pose and clock coordinates remain unknown. Each
product has full 62-dimensional interior and contains an isotropic
geometry ball of radius $.01\nu_r$ with the same motion and clock caps.
Table~\ref{tab:physical} gives physical units. The acceleration entries
are marginal bounds on one joint ball; they are not independently
saturable. The cone is a reference-instant acceleration-transient prior.

\begin{table}[b]
\centering\small
\caption{Physical radii/caps of the two certified products. Geometric and
rotation-vector bounds are per component; motion bounds are norms.}
\label{tab:physical}
\begin{tabular}{lrr}
\toprule
Quantity & Far & Near\\
\midrule
World coordinates [$\mu$m] & $.6$ & $20$\\
Mount coordinates [$\mu$m] & $.3$ & $10$\\
Rotation vector [$\mu$rad] & $.9$ & $30$\\
Speed [$\mu$m/s] & $2.4$ & $80$\\
Transl. acceleration [m/s$^2$] & $.0024$ & $.08$\\
Angular acceleration [rad/s$^2$] & $.0048$ & $.16$\\
Initial angular speed [rad/s] & $3.6\!\times\!10^{-5}$ & $.0012$\\
Independent clocks [ms] & $\pm6$ & $\pm6$\\
\bottomrule
\end{tabular}
\end{table}

Both products have a common positive-depth, simple-row branch. Their
complementary maps satisfy the global target bounds
\begin{equation}\label{eq:finite-secant}
 \|\Delta\bar v\|\leq C_r\|\Delta\mu_{+-}\|,
 \qquad C_F=20000,\quad C_N=270.
\end{equation}
The image norm includes all 96 raw coordinates, including noisy anchors.
Physical velocity has the extra factor $L/T=20$. These numerical constants
use $\Sigma=I$; for a fixed different SPD shape the whitened constants
may be multiplied by $\sqrt{\lambda_{\max}(\Sigma)}$.

\subsection{A complete uniform secant certificate}
Use anchor coordinates $Y_i$ and centres
$\widetilde b_s=b_s+\tau_s\bar v$. Write the mean as $G(x,\tau)$,
where $x$ contains the 60 nonclock coordinates. For arbitrary endpoints,
\begin{equation}\label{eq:lshape}
\begin{aligned}
\Delta G={}&G(x_1,\tau_1)-G(x_0,\tau_1)\\
&+G(x_0,\tau_1)-G(x_0,\tau_0).
\end{aligned}
\end{equation}
The first leg varies nonclock coordinates in a convex ambient product.
On the second, motion is fixed: if
$\alpha=\|(\bar a_0,B_0)\|>0$, both clock derivatives share
$c=(\bar a_0,B_0)/\alpha$ and
$w=\omega_0/\alpha$, with $\|w\|\leq .15$.
At $\alpha=0$ the cone forces $\omega_0=0$ and this clock leg vanishes
exactly. The construction does not divide by a realized zero amplitude.

An exact triangular shear retains finite clocks in the reference
linearization. Per auxiliary it changes the formal jet by
$\tau_s(\delta a,\delta B)$ and changes pose/centre by
\begin{equation}
\begin{aligned}
\delta\psi_s&\mathrel{+}=2\tau_s\delta\omega+\tau_s^2\delta B,\\
\delta\widetilde b_s&\mathrel{+}=\tfrac12\tau_s^2\delta a\\
&\quad+2[b_s]_\times(\tau_s\delta\omega+\tfrac12\tau_s^2\delta B).
\end{aligned}
\end{equation}
These identities hold as exact polynomials in the two independent clocks.

The anchor means are exactly
$(Y_{i1}/Z_i,Y_{i2}/Z_i)$ and
$(Y_{i1}/Z_i,(Y_{i2}-.2)/Z_i)$ for every motion in the chart.
Positive disparity gives an explicit inverse. The inverse image of a
line segment between actual anchor means is a positive weighted
interpolation of its endpoint $Y_i$ coordinates, so it remains in their
convex box. We retain the signed inverse Jacobian when eliminating
point differences; the 48 noisy anchor coordinates are not discarded.

For either fixture the remaining $48\times36$ rational augmented matrix
$A$ has rank 35. An exact $W$ satisfies
$WA=I-nn^T/\|n\|^2$. The two null jet blocks are
$(0,0,-.1,-1,1,0)$ and $(0,0,-.1,-1,-1,0)$.
Their combined squared residual outside $\operatorname{span}(c)$ is at
least 2 for every unit $c$. If $z=\eta+\lambda U_\tau^{-1}n$, then
\begin{equation}
 |\lambda|\leq\sum_{s,j}
 \frac{\sqrt{2.01}+|n_{sj}|}{4}|\eta_{sj}|.
\end{equation}
The common direction, rather than a direction grid, yields the uniform
kernel correction. It retains the nonzero velocity and angular-velocity
components of $n$.

We subtract the complete finite-clock linearization before outward
interval differentiation. This yields
\begin{equation}\label{eq:majorant}
 |z|\leq d\|\Delta\mu\|_2+M|z|,\qquad M\geq0.
\end{equation}
Positive weights verify $Mw\leq qw$, and a positive supersolution
verifies $d+Mh\leq h$. For the far fixture,
$q\leq .829645$ and $\|h_v\|_2<19550.992$.
The near fixture uses the correlated construction below and gives
$q\leq .506687$ and $\|h_v\|_2<269.363$.
These inequalities establish~\eqref{eq:finite-secant}; eigensolvers and
linear solves only propose candidates for the verified inequalities.

\paragraph{Preserving shared coefficients.}
For the near fixture we write the exact finite-clock expansion
\begin{equation}
 G=G_0+J_\tau\delta x+Q_\tau(\delta x,\delta x)+R_{3,\tau}.
\end{equation}
The quadratic coefficients are rational polynomials in the clocks. We
multiply their shared observation vectors by exact $W$ before taking
absolute bounds. Fixed-$Y$ partial derivatives remain fixed-$Y$
partials. Only the values of their state-dependent coefficients use
\begin{equation}\label{eq:Y-association}
\begin{aligned}
\delta Y_i={}&\delta P_i-x_i\bar v-\tfrac12x_i^2\bar a
 -2x_i[Y_{i0}]_\times\omega\\
&-x_i^2[Y_{i0}]_\times B+R_{Y,i},
\quad x_i=Y_{i0,1}/Y_{i0,3}.
\end{aligned}
\end{equation}
The code encloses $R_Y$ by the exact Cayley remainder and the change in
row, including all ordinary-world point deviations. The first leg of
\eqref{eq:lshape} preserves this affine association by endpoint
interpolation in convex boxes. On the second leg, Hadamard integration
halves direct motion deviations in the quadratic mixed derivative;
held-fixed endpoint $Y$ deviations retain their full coefficient.
Thus the motion contribution already present in the endpoint $Y$ is
not incorrectly halved.

We enclose $R_3$ with factored interval Taylor arithmetic. Products,
reciprocals and rotation series carry the full degree-at-least-three
remainder. Implicit row elimination is also exact: if the row shift
obeys $u=f_1+f_2(u)+R_f(u)$ and $b_\ell$ is the linear candidate-row
coefficient of the quadratic output, its cubic remainder is
$b_\ell(f_2(u)+R_f(u))+R_\ell(u)$.
Total implicit differentiation supplies every first and mixed
motion/clock second derivative. Root self-mapping, positive depth and
contraction are verified on the entire ambient product before these
bounds are used.

\paragraph{Ordinary-world inclusion.}
For far/near fixtures the anchor row caps are $.12/.23$.
The real Cayley bound $\|R-I\|\leq2\|k\|$ gives
$|Y_{ij}-P^r_{ij,0}|\leq .073328\nu_F$ and
$.057904\nu_N$, respectively, both below the used $.08\nu_r$ box.
The centre boxes also include the complete $\tau_s\bar v$ shift.
The near auxiliary depth and row-factor lower bounds exceed $.499876$
and $.999626$. Consequently these are certificates for ordinary-world
physical products, rather than replacement priors on triangulated points.

\subsection{Finite Gaussian risk constants}
For either fixture, the corrected repeated path with $y=.05$, $t=.1$
is inside the same product for
$\alpha\leq\bar\alpha_F=3\times10^{-6}$ or
$\alpha\leq\bar\alpha_N=10^{-4}$.
All 96 first-order mean differences vanish exactly. Interval second
derivatives along both endpoints give
\begin{equation}
 \|\Delta\mu_{++}\|\leq K_r\alpha^2,
 \quad K_F=.277094,\quad K_N=.581045.
\end{equation}
The physical target gap is $.2\alpha$\,m/s.
Choosing $\alpha^2=\min\{\bar\alpha_r^2,\sigma/K_r\}$ gives
\begin{equation}
\begin{aligned}
R^*_{++}&\geq\frac{.2^2}{16}
 \min\{\bar\alpha_r^2,\sigma/K_r\},\\
R^*_{+-}&\leq\min\{V_{\max,r}^2,
 4\cdot96\cdot20^2 C_r^2\sigma^2\}.
\end{aligned}
\end{equation}
The upper bound combines class-global least squares with the zero
predictor. On the unsaturated testing branch, the physical bounds are
\begin{equation}\label{eq:finite-risk}
\begin{array}{c|cc}
& R^*_{++}/\sigma\ \text{lower}&R^*_{+-}/\sigma^2\ \text{upper}\\
F&1250/138547&6.144\!\times\!10^{13}\\
N&500/116209&1.119744\!\times\!10^{10}.
\end{array}
\end{equation}
The far/near pair is unsaturated through
$1.91527\times10^{-9}/4.46242\times10^{-6}$\,px.
Strict ordering of these conservative full-class bounds is sufficient
below $1.12777\times10^{-13}/2.95102\times10^{-10}$\,px.
These quantify a fixed-class statistical guarantee; they do not specify
ordinary imaging noise tolerances.

\subsection{Reconstruction within the far certificate}
We use a fixed inner product $\mathcal Q_F\subset\mathcal B^F$ with all
62 unknowns. Every point/pose/mount/clock component uses its bound in
\eqref{eq:box} times a coordinate in $[-1,1]$.
Velocity uses an inscribed cube of side radius $.004\nu_F/\sqrt3$.
For $u\in[-4,4]^6$ and $r\in[-4,4]^3$, the motion chart is
\begin{equation}
 (\bar a,B)=\frac{.2\nu_F u}{\sqrt{1+\|u\|^2}},\qquad
 \omega=\frac{.15\|(\bar a,B)\|r}{\sqrt{1+\|r\|^2}}.
\end{equation}
This changes no observation or unknown count. Both layouts use the
same fixed set, with neither the direction nor the two-endpoint family
supplied to the estimator. The complete tested path lies inside it.

We use six noise levels from $10^{-11}$ to $10^{-8}$\,px and eight
independent repeats per endpoint. Noise is paired across layout and
$z=0/.02$ mounting interventions. The rule is
$\alpha=\min\{3\times10^{-6},\sqrt{\sigma_{\rm px}/(768K_F)}\}$;
only the final level saturates. Both layouts receive two fixed
truth-independent physical starts and at most 1000 evaluations per
start. We select the smaller residual, never the smaller velocity
error. Centered residuals use 80-decimal means; all 62 analytic
parameterized columns are independently precision-checked.
The selected physical row branch is shared and all returned states
remain in the exact ordinary-world product. The $z=.02$ control uses
its own forward-safe mounting set, not the lateral inverse certificate.

The final numerical rows are in \texttt{data/certified\_reconstruction.json}; scores include every selected output and numerical failure. Complementary fits terminate successfully in $192/192$
cases, versus $41/96$ lateral and $89/96$ depth-offset repeated fits.
Stationarity uses the projected, column-scaled gradient in the original
bounded coordinates, independently of a potentially saturated tanh
optimization chart. No selected output reaches a chart bound in this study.

\paragraph{Residual and target checks.}
Every selected image cost is below the realized truth-noise cost.
For lateral complementary fits, the secant certificate gives
\[
 \|\widehat V-V\|^2\leq(20C_F\sigma)^2(\sqrt{2c}+\|Z\|)^2,
 \]
where $c=\tfrac12\|(\mu(\widehat\theta)-\mu(\theta))/\sigma-Z\|^2$.

The largest squared-error/bound ratio is $3.375\times10^{-5}$.
The full far target radius is $2.4\,\mu$m/s. On this two-endpoint
sequence, zero prediction has mean loss $\|\Delta V\|^2/2$;
lateral repeated MSE is $1.157$--$3.353$ times that value.
Complementary-minus-repeated intervals are negative at all noise
levels in both mounting conditions (eight repeat clusters, keeping
the two endpoints together). We score complete cohorts; stationary
subgroups do not replace them, and numerical fits do not establish
minimax attainment.

\subsection{Relation to practical-noise reconstruction}
K.1--K.4 concern fixed small physical products. L.12 specifies a
separate practical-noise population; X bounds worst-case risk at $.1$\,px.
The small-domain inverse constant is not extrapolated to those larger sets.

\subsection{A reusable clock-jet criterion}
Here $v,a$ denote the normalized coefficients $\bar v,\bar a$.
Work in fixed normalized units and a smooth anchor chart with smooth inverse,
$\Phi_Y(\theta)=(x,\tau)$, where $x$ contains all $60$ nonclock coordinates,
$q=(a,B)\in\mathbb R^6$, and $\widetilde b_s=b_s+\tau_s v$.
The target coordinate $v$ is unchanged by $\Phi_Y$.
Let $G(x,\tau)\in\mathbb R^{96}$ retain all image coordinates, including
noisy metric anchors, and let $x_*$ be the static reference.
Assume one safe $C^2$ real scan-root branch on a convex ambient product
$X\times\mathcal T$, with $\mathcal T$ a fixed compact convex clock box;
all nonclock and radial-motion integration paths must lie there.
For every other nuisance value, $G$ is exactly clock independent when
$q=\omega=0$; admissible endpoints obey $\|\omega\|\le D\|q\|$.
Let $E:\mathbb R^{60}\to\mathbb R^{72}$ embed nonclock coordinates,
$J_s$ select formal six-dimensional jet block $s$, and $\pi_v$ select $v$.
The augmented matrix $A$ and continuous invertible shear $U_\tau$ satisfy
\[
 \begin{aligned}D_xG(x_*,\tau)&=AU_\tau E,\\ D_k\partial_{\tau_s}G(x_*,\xi)(c,w)&=AV_s(c,w,\xi),\end{aligned}
\quad k=(q,\omega).
\]
The columns $V_s$ are continuous on the safe compact set of
$\xi\in\mathcal T$, $\|c\|=1$, $\|w\|\le D$.
For every pair $\tau,\xi\in\mathcal T$, the cross-clock identities are
\[
 \begin{aligned}J_jU_\tau^{-1}V_s(c,w,\xi)&=\mathbf{1}_{j=s}c,\\ \pi_vU_\tau^{-1}V_s(c,w,\xi)&=0.\end{aligned}
\]
Put $V=[V_1,V_2]$, and bound the \emph{entire} nonclock correction by
$S_{\max}=\sup_{\tau,\xi,c,w}\|E^\top U_\tau^{-1}V(c,w,\xi)\|<\infty$.
Continuity and compactness give $u=\sup_\tau\|U_\tau^{-1}\|<\infty$.
These identities hold for the present polynomial shear $U=I+N$, $N^2=0$:
it adds $\tau_s q$ to jet $s$, $2\tau_s\omega+\tau_s^2B$ to pose, and
$\tau_s^2a/2+2[b_s]_\times(\tau_s\omega+\tau_s^2B/2)$ to centre.
The column $V_s$ has jet $s$ equal to $c$, pose $2(w+\xi_sc_B)$ and
centre $\xi_sc_a+2[b_s]_\times(w+\xi_sc_B)$, with zero motion components;
therefore $U_\tau^{-1}V_s=V_s$ even when $\tau\ne\xi$.

Define $\mathcal Z_c=\{z:J_sz=\gamma_sc,\ s=1,2\}$ and let $P_c$ project
orthogonally onto $\mathcal Z_c^\perp$.
For an orthonormal basis $N_\tau$ of $\ker(AU_\tau)$, suppose
\[
 \begin{gathered}\delta=\inf_{\tau\in\mathcal T,\,\|c\|=1}\sigma_{\min}(P_cN_\tau)>0,\\ \nu=\sigma_+\delta/u,\qquad \sigma_+=\sigma_{\operatorname{rank}(A)}(A).\end{gathered}
\]
If $A$ has full column rank, set $\delta=1$.
The kernel projector is continuous; hence transversality
$\ker(AU_\tau)\cap\mathcal Z_c=\{0\}$ for every $\tau,c$ is equivalent
to a positive uniform angle. For $z\in\mathcal Z_c$, projection off that
kernel and the singular-value product inequality give
$\|AU_\tau z\|\ge\nu\|z\|$.
When $\ker A=\operatorname{span}\{n\}$, set $n_s(\tau)=J_sU_\tau^{-1}n$
and $H_\tau=\sum_s n_s(\tau)n_s(\tau)^\top$. Then
\[
 \delta^2=\inf_\tau
 \frac{\operatorname{tr}H_\tau-\lambda_{\max}(H_\tau)}{\|U_\tau^{-1}n\|^2}.
\]
For two clocks the numerator is $\lambda_2(H_\tau)$, positive exactly for
two nonzero, noncollinear transported null jets. If $n_q=0$, these jets
are independent of $\tau$ in the stated shear.

\paragraph{Finite-domain consequence and proof.}
For arbitrary endpoints split $\Delta G$ into the nonclock leg from $x_0$
to $x_1$ at $\tau_1$, and the clock leg from $\tau_0$ to $\tau_1$ at $x_0$.
If $\alpha=\|q_0\|>0$, put $c=q_0/\alpha$, $w=\omega_0/\alpha$ and
$\gamma=\alpha(\tau_1-\tau_0)$.
Let $B,H$ be the exact averaged nonclock and divided-clock matrices and
$\bar V$ the reference clock columns averaged along the clock leg.
Suppose a uniform whole-domain certificate gives
\[
 \|[B-AU_{\tau_1}E,\ H-A\bar V]\|\le\varepsilon,
 \qquad \varepsilon(1+S_{\max})<\nu.
\]
The exact split is $\Delta G=AU_{\tau_1}z+e$, where
$z=E\Delta x+U_{\tau_1}^{-1}\bar V\gamma\in\mathcal Z_c$,
$\pi_vz=\Delta v$, and $\|e\|\le\varepsilon\|(\Delta x,\gamma)\|$.
Since $\|\gamma\|=\|z_{\rm jet}\|$ and
$\Delta x=z_{\rm nc}-\bar S\gamma$, the inverse block shear gives
$\|(\Delta x,\gamma)\|\le(1+S_{\max})\|z\|$. Consequently
\[
 \|\Delta v\|\le\frac{\|\Delta G\|}{\nu-\varepsilon(1+S_{\max})}.
\]
If $\alpha=0$, the cone gives $\omega_0=0$ and exact clock invariance
removes the clock leg; take $\gamma=0$ and any unit $c$.
The endpoints can have different acceleration directions. On nested safe
products tending to $x_*$, the $C^2$ Hadamard formula and bounded $(c,w)$
give $\varepsilon\to0$ for a fixed clock box. Ordinary-world application
also requires chart inclusion and a uniform remainder bound. This
argument concerns velocity and allows the raw clocks to be
unidentifiable at $q=0$.

For repeated quarter turns, the null jets are collinear and
$n_v=2ye_3$. Appendix C constructs a finite witness for $y\ne0$.
At $y=0$ that null has zero velocity component.

\paragraph{Supplying one auxiliary clock.}
For the repeated $++$ layout, fixing either auxiliary clock forces its
jet block to vanish in every physical secant. Both transported null
jets are nonzero. The zero-jet subspace therefore intersects the
transported kernel trivially, and compactness gives a uniform
positive angle. The same remainder argument and class-global least
squares give $O(\sigma^2)$ normalized velocity risk on a sufficiently
small local class with that clock supplied. The rational static
66-column rank check is in \texttt{static/verify\_supplied\_aux\_clock.py}.

\nobalance
\subsection{Computational witnesses}
The archive README lists the exact reconstruction commands for
A/J (static ranks), C.1 (known-path dual), D/E (Gram and LDL objects),
and K.1--K.3 (complete Far/Near interval chains). The latter require
contraction below one and velocity constants below $20000/270$.
They regenerate mathematical witnesses rather than accepting saved
success flags.

\subsection{Optimization-free confidence-set witnesses}
We reuse the exact observations and prescribed pairs of K.4, without
new worlds, noise draws or fits. For $Y=\mu(\theta)+\sigma Z$,
$Z\sim N(0,I_{96})$, define
\[
\mathcal C(Y)=\{\theta\in\mathcal B:
\|\mu(\theta)-Y\|^2\leq\sigma^2\chi^2_{96}(.95)\}.
\]
Every admissible truth lies in this set with probability $.95$.
If both prescribed endpoints pass this image-residual test,
$\operatorname{diam}V(\mathcal C(Y))\geq\|\Delta V\|$ regardless
of any optimizer output. We evaluate both candidates with 80-digit
means and regenerate the existing Gaussian draws from the fixed seed.
The numerical cutoff is $\chi^2_{96}(.95)=119.8709393$.

Across the six levels, $++$ admits both endpoints in $90/96$ observations, versus $0/96$ for $+-$ and depth-offset $++$. These witnesses do not depend on optimizer convergence. On its certified domain the complementary secant bounds the full diameter by $2(L/T)C\sigma\sqrt{\chi^2_{96}(.95)}$. The final confidence-set counts are included in \texttt{data/certified\_confidence.json}.

\renewcommand{\theequation}{L.\arabic{equation}}
\renewcommand{\thetable}{L.\arabic{table}}
\renewcommand{\thefigure}{L.\arabic{figure}}
\setcounter{equation}{0}\setcounter{table}{0}\setcounter{figure}{0}
\section{Independent-scene reconstruction and timing cost}\label{app:L}

\subsection{Shared reconstruction protocol}
We use the implicit-row model with $L=1$\,m, $T=.05$\,s,
$f=768$\,px and independent $.1$-px Gaussian coordinate noise.
Both metric anchors remain noisy. Intrinsics, readout laws and relative
anchor calibration are supplied; points, motion, auxiliary poses and
clocks are unknown. The resource count is $8n$ measurements and
$3n+26$ unknowns: $384/170$ for $n=48$, or $96/62$ for $n=12$.
Worlds and noise are paired across layouts.

We initialize structure from noisy anchors, mounts from their nominal
values, and velocity/clocks at zero. Points stay within $.5$\,m per
coordinate of that initializer; mount coordinates within $.04$\,m of
nominal; rotation-vector coordinates within $.08$\,rad; and velocity
components within $.4619$\,m/s. For normalized $c=(a,B)$,
\[
 \begin{aligned}
 c&=.3r/\sqrt{1+\|r\|^2},&r&\in[-4,4]^6,\\
 \omega&=.15\|c\|u/\sqrt{1+\|u\|^2},&u&\in[-4,4]^3.
\end{aligned}
\]
Fitted clocks lie within $\pm6$\,ms. Two truth-independent starts have
opposite nonzero angular-acceleration initializations. We use bounded
TRF with Jacobian scaling, 300 evaluations per start and stopping
tolerances $(\texttt{ftol},\texttt{xtol},\texttt{gtol})=(10^{-12},10^{-11},10^{-10})$,
selecting by image residual. We retain every selected return. Unless stated otherwise,
loss is $\min(\|\widehat V-V\|^2,1)$, with unavailable returns assigned 1.
No cap activates in the reported primary conditions. A projected,
column-scaled gradient of $10^{-5}$ diagnoses stationarity.

\subsection{48-point accuracy and constant predictors}
Each of two disjoint cohorts has 32 worlds, sixteen per transient
family. Points have depth 2--5\,m and normalized lateral coordinates
within $\pm.22$; nominal mounts are $(\pm.12,.05,0)$\,m with
independent $\pm2$-mm perturbations, rotation-vector perturbations
$\pm.006$\,rad, and true clocks within $\pm4$\,ms.
The mean-shifted velocity is $(.08,-.06,.18)$\,m/s plus independent
$\pm.04$-m/s component perturbations. The isotropic cohort uses uniform
sphere directions and speeds uniform on $[.16,.4]$\,m/s.
Both use the two motion laws specified in L.12.

In $++/--/+-$ order, RMSE is $.079777/.084173/.059704$\,m/s
(mean-shifted) and $.087501/.080627/.059164$\,m/s (isotropic).
We use 10,000 whole-world bootstrap draws within the sixteen-world
motion strata; the two primary contrasts per cohort have nominal
97.5\% intervals (Bonferroni 95\% within that cohort).

\begin{table}[H]
\centering\small
\caption{Complementary-minus-repeated mean squared loss and paired
97.5\% world-bootstrap intervals, in $10^{-3}$\,(m/s)$^2$.
Each row uses all 32 worlds in its cohort.}
\label{tab:Lcontrasts}
\begin{tabular}{llrr}
\toprule
Cohort & Against & Difference & Interval\\
\midrule
Mean-shifted & $++$ & $-2.800$ & $[-5.888,-.397]$\\
 & $--$ & $-3.520$ & $[-5.734,-1.588]$\\
Isotropic & $++$ & $-4.156$ & $[-7.998,-.932]$\\
 & $--$ & $-3.000$ & $[-5.816,-.491]$\\
\bottomrule
\end{tabular}
\end{table}

Complementary errors are below $.1$\,m/s in $30/32$ and $31/32$ worlds.
Zero RMSE is $.209741/.289188$\,m/s. The shifted population's known
mean gives $.041536$\,m/s; zero is the optimal population constant
for the isotropic replication.

\subsection{Supplied-clock control on the isotropic cohort}
Supplying both true clocks leaves 168 unknowns and changes no other
input or setting. All 96 fits are stationary; RMSE becomes
$.069185/.071547/.058016$\,m/s. Unknown-minus-known mean squared losses
are $.002870/.001382/.000135$\,(m/s)$^2$.
Complementary-minus-repeated timing penalties are $-.002735/-.001247$,
with 97.5\% stratified-bootstrap intervals
$[-.005855,-.000196]/[-.002317,-.000255]$ (100,000 draws).
The supplied-clock layout contrasts themselves have descriptive
intervals crossing zero.

\subsection{Independent angular-rate bounds}
On the same isotropic worlds, we replace the fitted cone by independent
physical angular-rate component bounds $\pm1.6$\,rad/s. We also test
$\omega'=\omega-.0125u$, $u\sim\mathrm{Unif}(S^2)$, adding a
$.5$-rad/s reference-rate vector. Geometry, other motion coefficients,
clocks and Gaussian draws stay paired; perturbed means are regenerated.
Reference-rate norms become $.235$--$.836$\,rad/s and 14/32 truths
leave the original cone. All 384 selected returns are stationary.

\begin{table}[H]
\centering\small
\caption{Complete rate/prior diagnostic with independent angular-rate
bounds in all rows. Each row uses the same 32 worlds; rate is the
added reference angular-rate magnitude [rad/s]. RMSE is in m/s.
All returns are finite; these are not new independent cohorts.}
\label{tab:Lrate}
\begin{tabular}{llrrr}
\toprule
Rate & Clocks & $++$ & $--$ & $+-$\\
\midrule
0 & Unknown & .08763 & .08063 & .05916\\
0 & Known & .06919 & .07155 & .05802\\
.5 & Unknown & .08755 & .08085 & .05914\\
.5 & Known & .06927 & .07192 & .05808\\
\bottomrule
\end{tabular}
\end{table}

For perturbed rate and unknown clocks, complementary excess mean losses
are $-.004167/-.003038$\,(m/s)$^2$, with 97.5\% stratified-bootstrap
intervals $[-.008010,-.000852]/[-.005778,-.000544]$ (100,000 draws).
Errors are below $.1$\,m/s in 31/32 worlds versus 24/32 for either repeat.

\subsection{Finite physical pairs at $.1$\,px}\label{sec:pair-tests}
We evaluate Appendix C's corrected path on the twelve rational points,
using separate paths for $++$ and $--$ and comparing each fixed pair
under repeated and complementary readout. The prescribed grid is
$\delta_V\in\{.01,.025,.05,.1\}$\,m/s,
$y\in\{.02,.05,.1,.15,.2\}$\,m, $z\in\{0,.02\}$\,m and clock steps
$\{2,4,6,10,15,20\}$\,ms. Set $\alpha=\delta_V/(40yt)\leq.2$.
For each fixed $(y,z)$, both states must lie in the same physical class:
\[
\begin{aligned}
 |P_{ij}-P_{ij,0}|&\leq.5,& |v_j|&\leq.04/\sqrt3,\\
 \|(a,B)\|&\leq.3,& |\omega_j|&\leq.04,\\
 |\psi_{sj}|&\leq.08,& |b_{sj}-b_{sj,0}|&\leq.04,\quad |\tau_s|\leq.4.
\end{aligned}
\]
Of 480 grid cases, 344 pass the parameter bounds; all have positive
depth, row factor at least $.5$, $|u|\leq.5$ and row residual below
$10^{-12}$. These classes differ from Appendix K's certified products.

For two Gaussian means, $d=\|\Delta\mu\|/\sigma$ gives optimal
balanced decision error $p_e=\Phi(-d/2)$. Forty-seven cases satisfy
$\delta_V\geq.05$\,m/s, repeated $p_e\geq.25$ and complementary
$p_e\leq.05$. We report the passing zero-depth case with smallest
repeated distance for each sign/gap and its matched depth control.

\begin{table}[H]
\centering\small
\caption{Selected finite pairs and matched depth controls at $.1$-pixel
noise. $d_R,d_C$ are image distances in noise-standard-deviation units;
$p_R$ is repeated-readout balanced decision error. Complementary error
is below $10^{-100}$ in every row. Each same-state layout comparison
retains all 96 observations.}
\label{tab:practicalpairs}
\begin{tabular}{lrrrrr}
\toprule
Repeat & $\delta_V$ [m/s] & $z$ [m] & $d_R$ & $d_C$ & $p_R$\\
\midrule
$++$ & .05 & 0 & .47720 & 44.7528 & .40571\\
$++$ & .05 & .02 & .58066 & 45.1640 & .38578\\
$--$ & .05 & 0 & .29552 & 45.0675 & .44127\\
$--$ & .05 & .02 & .30818 & 45.6315 & .43877\\
$--$ & .10 & 0 & 1.17897 & 90.1598 & .27777\\
$--$ & .10 & .02 & 1.14298 & 91.2785 & .28383\\
\bottomrule
\end{tabular}
\end{table}

All reported pairs use $y=.2$\,m and a 4-ms clock step. Their amplitudes
are $.078125/.15625$ at gaps $.05/.1$\,m/s; reference translational
accelerations are $12.5/25$\,m/s$^2$ and angular accelerations
$88.4/176.8$\,rad/s$^2$. Independent 80/120-digit projection checks agree.
For endpoint posterior weight $q$, squared Bayes loss is
$\delta_V^2q(1-q)\geq\delta_V^2\min(q,1-q)/2$. Hence
\[
 \sup_{\theta\in\{\theta_0,\theta_1\}}
 \mathbb E_\theta\|\widehat V-V(\theta)\|^2\geq\delta_V^2p_e/2.
\]
The $.05$-m/s zero-depth pairs give root-risk lower bounds
$.02252/.02349$\,m/s. Discriminating these supplied pairs is distinct
from reconstructing unknown worlds or bounding whole-class risk.

\subsection{Acceleration and timing controls}
We use 32 independent 48-point worlds, reference angular speed
$.5$\,rad/s, and the independent rate bounds of L.4. Multiplying both
$a,B$ by $\lambda\in\{0,.1,.3,1\}$ retains geometry, velocity,
clocks and noise. Unknown/supplied-clock fits retain 170/168 unknowns;
all other settings follow L.1. All 768 selected returns are stationary. Numerical rows for main Fig.~5 accompany the archive in \texttt{data/figure5\_numerical\_rows.csv}.
At $\lambda=1$, translational/angular acceleration norms span
$1.674$--$45.131$\,m/s$^2$ and $16.230$--$120.380$\,rad/s$^2$;
other levels scale these ranges. Cayley motion specifies the trajectory,
not constant angular speed throughout the exposure.

At zero acceleration and unknown clocks, complementary excess losses
are $+.00024045/-.00175302$\,(m/s)$^2$, with nominal 97.5\% intervals
$[-.00312286,+.00358822]/[-.00544889,+.00170671]$ from 100,000
stratified world-bootstrap draws. Both cross zero. At $.1/.3$, the
marginal unknown-clock intervals are negative; supplied-clock layout
intervals cross zero at every level. Clock-bound coordinate counts
(out of 64) for $++/--/+-$ are $33/32/26$, $24/17/9$, $15/5/0$ and
$8/0/0$ as $\lambda$ increases. The primary family is the two zero-level
contrasts; other comparisons are descriptive.

\subsection{Independent twelve-point inputs}
We generate 32 further worlds with twelve noncoplanar points at 2--5\,m,
isotropic velocity directions, speeds uniform on $[.16,.4]$\,m/s,
sixteen worlds per motion family, and true clocks within $\pm4$\,ms.
An independent scalar generator uses SciPy auxiliary rotations, direct
Cayley solves and bracketed Brent row roots. Observations have paired
$.1$-px noise; all 62 parameters remain unknown. Fits follow L.1.

RMSE is $.34816/.37634/.21250$\,m/s, versus $.29076$ for zero.
Table~\ref{tab:originalresource} uses 100,000 whole-world bootstrap draws
within motion strata. All 96 selected returns are finite and uncapped;
one $++$ return exceeds the stationarity threshold
($1.32\times10^{-4}$) and remains in the scores.

\begin{table}[H]
\centering\small
\caption{Independent-input twelve-point contrasts in (m/s)$^2$.
Negative differences favor complementary readout. The two primary
97.5\% intervals use a Bonferroni nominal 95\% joint construction;
the zero-predictor comparison is a separate secondary diagnostic.}
\label{tab:originalresource}
\begin{tabular}{lrr}
\toprule
Against & Mean & Interval\\
\midrule
$++$ (primary) & $-.076059$ & $[-.118484,-.036947]$\\
$--$ (primary) & $-.096476$ & $[-.182128,-.037756]$\\
Zero (secondary) & $-.039383$ & $[-.057804,-.020675]$\\
\bottomrule
\end{tabular}
\end{table}

\subsection{Scoring conventions}
Worlds are the independent units. Bootstrap intervals have nominal
coverage; L.12 instead supplies finite-sample confidence bounds.

\subsection{Expanded optimization on the same inputs}
On L.7's inputs, we add six 1200-evaluation starts to the two original
300-evaluation returns: the two original initial states and four
truth-independent random states, paired across layouts. Minimum-residual
selection among all eight gives RMSE $.34374/.37971/.21250$\,m/s.
All 96 selected returns are stationary. Complementary excess losses are
$-.073000/-.099024$\,(m/s)$^2$, with nominal 97.5\% intervals
$[-.118745,-.030006]/[-.185168,-.041507]$ (100,000 stratified
world-bootstrap draws). Secondary changes in the layout gap have
intervals crossing zero.

\subsection{Source-selected ambiguities}\label{sec:source-pairs}
Using L.7's 32 worlds, we fix axial velocity at nine equally spaced
values on $[-.46188,.46188]$\,m/s and fit the other 61 coordinates.
Each query uses the anchor-based start and L.9's selected state,
with the queried coordinate replaced, and a 600-evaluation cap.
No truth or noise vector enters this search. We require the common
physical envelope: point $x,y\in[-1.5,1.5]$\,m and $z\in[1.5,5.5]$\,m;
$|\bar v_j|\leq.04/\sqrt3$, $\|(\bar a,B)\|\leq.3$,
$\|\omega\|\leq.15\|(\bar a,B)\|$; mount/rotation perturbations
$.04$\,m/$.08$\,rad and clocks $\pm6$\,ms.
All 864 queries return valid candidates inside the fixed cutoff
$\|\mu-Y\|^2/(.1/768)^2<\chi^2_{96}(.95)-10^{-6}$,
$\chi^2_{96}(.95)=119.870939$.

At each velocity retain the smaller independently verified residual.
For source layout $R$, select the closest of its 36 pairs with gap
$\delta\geq.05$\,m/s, breaking ties by grid indices. Keep both complete
states fixed under other readouts. With $p_d=\Phi(-d_d/2)$, the criterion
is $p_R\geq.25$ and target $p_d\leq.05$.

\begin{table}[H]
\centering\small
\caption{Source-selected known-pair tests at $.1$\,px. The common
criterion is $\delta\ge .05$\,m/s, $p_R\ge .25$ and
$p_{+-}\le .05$. All 32 reused worlds remain in each denominator.
Ranges cover every selected pair; these are descriptive counts.}
\label{tab:compatibility-pairs}
\begin{tabular}{lrr}
\toprule
Source layout $R$ & $++$ & $--$\\
\midrule
Criterion met & $18/32$ & $20/32$\\
Valid, criterion not met & $14/32$ & $12/32$\\
Missing/unknown pair & $0/32$ & $0/32$\\
Velocity gap [m/s] & $.11557$--$.16904$ & $.11547$--$.18899$\\
$p_R$ range & $.32479$--$.46715$ & $.29058$--$.47670$\\
Median $p_{+-}$ & $.04327$ & $.03092$\\
\bottomrule
\end{tabular}
\end{table}

Selecting on $+-$ instead gives $p_{+-}\geq.25$ in 31/32 worlds and
meets the cross-readout criterion in $10/32$ ($++$ target) and $11/32$
($--$ target), as in main Fig.~3. Switching repeated-source pairs to
the opposite repeat meets the criterion in $27/32$ and $29/32$ worlds;
its error is lower than $+-$ on all 64 pairs. These are selected-pair
comparisons, not whole-class risk bounds.

Pooling the same 27 candidates per world gives 11,211 eligible pairs.
Each layout's hardest pair comes from its own fits in all 96 cases;
median maximum errors are $.412/.416/.337$. Seven pooled states fit
all three observations and yield eligible joint pairs in two worlds,
both hardest under $+-$. Appendix W examines compatibility diameters
on these same observations.

\raggedbottom
\subsection{Geometry conditioning at finite noise}
\label{sec:original-finite-noise}
\label{sec:conditioning}
Table~1 imposes quantitative geometry bounds; the broader generator allows arbitrarily small point spread.

\paragraph{An original-contract obstruction.}
Let $\mathcal S$ denote the common valid physical part of the original
twelve-point generator support, with positive, simple, selected row
branches under all three layouts and the original motion bounds.
The near-hard family's unbounded Gaussian perturbations are not assumed
to define valid observations everywhere. Point depths are in $[2,5]$\,m, normalized lateral
coordinates in $[-.22,.22]$, and physical speeds in $[.16,.4]$\,m/s.
The original auxiliary poses, mounts, clocks and motion coefficients
remain unknown. For every positive noise and every layout,
\begin{equation}\label{eq:support-risk}
 \inf_{\widehat V}\sup_{\theta\in\mathcal S}
 \mathbb E_\theta\|\widehat V-V(\theta)\|^2
 =(.4\,\mathrm{m/s})^2.
\end{equation}
The same value holds when coincident/coplanar configurations are excluded
but arbitrarily close distinct noncoplanar configurations remain allowed.
This is a support worst-case result, not the risk averaged over the
generator distribution or the finite cohort.

\paragraph{Construction and safe branch.}
Use the original normalized units $L=1$\,m, $T=.05$\,s and nominal
auxiliary centres $(\pm.12,.05,0)$, with zero pose perturbations.
First put all twelve points at $p=(0,0,3)$ and choose
\[
 a=.1e_3,\quad B=0,\quad\omega=.005e_3,\quad v_\pm=\pm.02e_3.
\]
These are generic-family support values:
$\|(a,B)\|=.1$ and $\|\omega\|/\|(a,B)\|=.05$.
Physical velocities are $V_\pm=\pm.4e_3$\,m/s. Rotation is about $e_3$,
so $R(t)[p-C(t)]$ remains axial. Both anchor scan rows are zero.
Each auxiliary's static scan row is $r_s=1/60$ for $Q_+$ and
$r_s=-1/60$ for $Q_-$. Set $\tau_s=-r_s$, only $\mp.8333$\,ms
physically, within the original true-clock range. The clocks remain
fixed between the two states. At every camera's root $x_c=r_c$,
the exposure time is exactly zero. Consequently every camera sees
$Q_c(p-b_c)$, independently of $v_+$ versus $v_-$.

For either state the auxiliary scan map is
$n_s/[3-v_z t-.05t^2]$, with $n_s=\pm.05$.
On the whole row interval $[-.6,.6]$, $|t|\leq37/60$,
depth exceeds $2.968$ and the scan derivative has magnitude below
$.00047$. The map takes this interval strictly into itself and is a
contraction, so the root is unique with signed row factor above
$.99953$. The anchor scan map is identically zero. Thus the construction
uses the selected safe branch, with genuinely nonzero acceleration
and angular velocity.

\paragraph{Sharp lower bound with noncoplanar points.}
Replace $p$ by $P_i(\eta)=p+\eta q_i$ for twelve fixed distinct offsets
whose first four form a nonzero-volume tetrahedron. For sufficiently
small $\eta>0$ these points lie in the original generator support, are
noncoplanar, and retain the strict branch margins. Both complete mean
vectors converge to the same vector as $\eta\to0$.
For a Gaussian two-state experiment with velocity gap $\Delta$ and
noise-normalized mean distance $d$, the equal-prior Bayes squared risk
is at least
\begin{equation}\label{eq:sharp-pair-lower}
 \frac{\Delta^2}{2}\Phi(-d/2).
\end{equation}
Indeed, its exact risk is
$\frac{\Delta^2}{2}\int f_0f_1/(f_0+f_1)$; use
$f_0f_1/(f_0+f_1)\geq\min(f_0,f_1)/2$ and
$\int\min(f_0,f_1)=2\Phi(-d/2)$.
Here $\Delta=.8$\,m/s and $d\to0$ at every fixed positive noise, so
the supremum risk is at least $.16$\,(m/s)$^2$. The zero estimator
has risk at most $.16$ on the whole speed-bounded support, proving
Eq.~\eqref{eq:support-risk}. Boundary speeds and shell values can also
be approached from the interior. Each layout has its own admissible
least-favorable clocks in the same containing physical support.

The broader fitting envelope permits velocity components within
$\pm.8/\sqrt3\simeq\pm.46188$\,m/s. Taking opposite cube corners, and zero quadratic
and angular motion, gives the analogous sharp root risk $.8$\,m/s.
This envelope and the radius-$.4$ generator support are different classes.
Neither result applies to Table~1's separated rational-reference products.

\paragraph{Deterministic replay at $.1$\,px.}
We use twelve fixed integer offsets beginning with $0,e_1,e_2,e_3$;
the complete list is in the packaged script. We retain all three layouts
at each prescribed $\eta$.
Two independent existing forward implementations agree within
$10^{-12}$ in normalized image coordinates. At $\eta=.1/.01/.001$\,m,
all layouts give root-risk lower bounds above $.2695/.3921/.3992$\,m/s;
as $\eta$ decreases, all approach $.4$\,m/s. These are constructed
geometry checks, not random reconstruction trials. The packaged
\texttt{verify\_obstruction.py} and \texttt{OBSTRUCTION\_REPLAY.json}
retain complete states, image distances, signed branch diagnostics and
all six geometry scales.

L.12 instead bounds distribution-average reconstruction risk.

\subsection{A nonasymptotic practical-noise risk comparison}
\label{sec:finite-risk-confirmation}
We certify distribution-average error for the original 96-observation,
62-unknown estimator at $.1$\,px.

\paragraph{Population and physical scales.}
We use $L=1$\,m, $T=.05$\,s and $f=768$\,px, with
$\Sigma=I_{96}$ and the row law $t=x+\tau_c$. All coefficients below are normalized; $V=20\bar v$\,m/s. The two motion laws share scene, velocity and mounting draws.

For $i=1,\ldots,12$, draw $z_i\sim U[2,5]$ and
$\xi_i,\eta_i\sim U[-.22,.22]$, and set
$\bar P_i=z_i(\xi_i,\eta_i,1)^T$.
Draw $\bar v=s u_v$, where $s\sim U[.008,.020]$ and
$u_v$ is uniform on the unit sphere $S^2$. Thus physical speeds
are uniform on $[.16,.4]$\,m/s with isotropic direction.
Anchor centres are $0$ and $.2e_2$, with identity rotations and zero
clock offsets. Each auxiliary centre independently perturbs
$(.12,.05,0)$ or $(-.12,.05,0)$ by componentwise $U[-.002,.002]$.
Auxiliary rotation-vector components are $U[-.006,.006]$ and enter
$Q_s=Q_{\mathrm{nom},s}\exp([\psi_s]_\times)$.
Each auxiliary clock is $\tau_s\sim U[-.08,.08]$, corresponding to
$\pm4$\,ms. Nominal rotations change with the layout as in the main model.

\paragraph{Near-hard motion law.}
Draw $\alpha\sim U[.035,.11]$, $h\sim U[.035,.095]$ and
independent $g_a,g_B,g_\omega\sim N(0,I_3)$. Set
\begin{equation}\label{eq:confirmation-near-hard}
\begin{aligned}
 \bar a&=-.1\alpha e_3+.0007g_a,\\
 B&=\alpha(-1,1,0)^T+.003g_B,\\
 \omega&=\alpha h e_1+.0003g_\omega.
\end{aligned}
\end{equation}
The center follows the $++$ compensation direction; the Gaussian perturbations have unbounded raw support before conditioning.

\paragraph{Random-direction motion law.}
Independently draw $u_6$ uniform on $S^5$, $u_\omega$ uniform on $S^2$,
$r\sim U[.05,.15]$ and $\gamma\sim U[.025,.09]$. Set
\begin{equation}\label{eq:confirmation-random-direction}
 q=(\bar a,B)=r u_6,\qquad \omega=\gamma r u_\omega.
\end{equation}
The released data label this motion family \texttt{generic}; it does
not denote generic mounting. All primitive uniforms, sphere directions
and Gaussians are mutually independent before the stated transformations
and conditioning, and worlds are independent.

\paragraph{Conditioning, mixture and noise.}
Let $\mathcal E$ be the common truth-only event that
$\|q\|<.3$, $\|\omega\|<.15\|q\|$, and every observation in
each of $++/--/+-$ has its selected row root in $[-.5,.5]$ with
positive depth and positive row factor. The implemented event also
requires the fitting forward map to evaluate successfully and agree
with the independent projection to absolute error $<10^{-11}$ in every normalized
image coordinate. This deterministic check uses no noise or fitted
output. The target is
\begin{equation}\label{eq:confirmation-population}
 \mathsf P=\tfrac12\mathsf P_{\mathrm{near}}(\,\cdot\mid\mathcal E)
          +\tfrac12\mathsf P_{\mathrm{random}}(\,\cdot\mid\mathcal E).
\end{equation}
We draw 128 independent worlds per conditional law, without a spread
filter. Truth-only joint gates precede fitting; any failure stops the
cohort without replacement. All 256 pass. Conditional on completion,
the product law is separately conditioned by family, as above.

For each world draw $Z\sim N(0,I_{96})$ independently of the truth and
other worlds, and supply
$Y_d=\mu_d(\theta)+(.1/768)Z$ to layout $d$. The same truth and $Z$
are paired across its three layouts. No realized nuisance parameter is supplied. The archive provides the independent input implementation.

\paragraph{Fixed estimator.}
The estimator receives observations and layout only. All 62 coordinates
remain unknown. It uses the original two starts, 300 evaluations per
start, bounds, tolerances and residual selection. If initialization fails or neither start returns a finite estimate,
the output velocity is zero. Every finite
return is retained regardless of numerical termination or stationarity.
True speed is at most $.4$\,m/s and the fitted velocity cube has norm at
most $.8$\,m/s, so uncapped squared loss $L_d\leq1.44$\,(m/s)$^2$.
We fix the conservative bound $b=1.45$. No global optimization
assumption is needed.

\paragraph{Finite-sample upper confidence bound.}
We use the e-value averaging principle of Vovk--Wang [24], with the
following self-contained bounded-loss construction. For $n$ independent,
possibly nonidentically distributed values
$D_i\in[-b,b]$, let $m=n^{-1}\sum_i\mathbb E D_i$. Here $D_i$ is either
a paired loss difference $L_{+-,i}-L_{R,i}$ or $L_{+-,i}$ itself.
For $\mu<b$ and fixed $\lambda\in\Lambda=\{1/4,1/2,3/4,1\}$ define
\begin{equation}\label{eq:risk-evalue}
 E(\mu)=\frac14\sum_{\lambda\in\Lambda}
 \prod_{i=1}^n\left[1+\lambda\frac{\mu-D_i}{b-\mu}\right].
\end{equation}
Every factor is nonnegative. If $m\geq\mu$, independence and the
arithmetic--geometric mean inequality give
\begin{equation}\label{eq:evalue-validity}
\begin{split}
 \mathbb E\prod_i\left[1+\lambda\frac{\mu-D_i}{b-\mu}\right]
 &=\prod_i\left[1+\lambda\frac{\mu-\mathbb E D_i}{b-\mu}\right]\\
 &\leq\left[1+\lambda\frac{\mu-m}{b-\mu}\right]^n\leq1.
\end{split}
\end{equation}
Therefore $\mathbb E E(\mu)\leq1$ under this null. Each sample factor
is nondecreasing in $\mu$, since its derivative is
$\lambda(b-D_i)/(b-\mu)^2\geq0$.
Inverting $E(U)=1/\delta$ gives an upper confidence endpoint with
$\Pr\{m>U\}\leq\delta$, by Markov's inequality at the fixed true $m$.
The limiting endpoint is $b$ if no crossing exists.
Unequal family-specific expected losses are allowed: only the average
of the 128+128 expectations enters the null. Pairing within a world
does not require independence of the three layouts; independence is
used across worlds.

We prespecify three statements: the two complementary-minus-repeated
mean losses and the complementary mean loss itself, each with
$\delta=.05/3$. Bonferroni gives at least $95\%$ simultaneous coverage.
We freeze the lambda mixture, sample count, inputs and solver before
confirmation. These finite-sample bounds concern expected uncapped
reconstruction loss under the stated simulated population.

\paragraph{Confirmation results.}
We evaluate all 256 sampled worlds in 768 conditions. Both starts
return finite estimates in every condition. No condition is replaced or omitted, and no
fallback is used. Complementary RMSE is $.2267462$\,m/s, versus
$.3562273/.3538932$\,m/s for $++/--$. Mean uncapped squared losses are
$.05141384/.12689787/.12524039$\,(m/s)$^2$ for $+-/++/--$.
The three simultaneous upper confidence endpoints are
\begin{equation}\label{eq:finite-risk-results}
\begin{aligned}
 \mathbb E(L_{+-}-L_{++})&\leq-.03987986,\\
 \mathbb E(L_{+-}-L_{--})&\leq-.03892862,\\
 \mathbb E L_{+-}&\leq .08067585
\end{aligned}\qquad[(\mathrm{m/s})^2].
\end{equation}
These $95\%$ simultaneous bounds give root expected squared error
below $.28404$\,m/s for $+-$ under the conditional mixture.
Independent replay checks all losses and endpoints. All selected fits
stop before the budget and meet the $10^{-5}$ projected-gradient threshold.
Clock/velocity bound counts are 151/164, with no structure bound active.

\paragraph{Common speed-support control.}
Projection onto the $.4$\,m/s speed ball preserves the absolute
risk bound, since it cannot increase squared loss for admissible truth.
In a post-hoc family of four contrasts we use $\delta=.05/4$,
$b=1.45$ for raw loss minus zero, and $b=.65$ for projected
contrasts. The simultaneous $95\%$ upper endpoints for $+-$
versus zero are $+.00649$ raw and $-.01874$ projected.
Projected $+-$ versus projected $++$ and $--$ gives
$-.02511$ and $-.02502$\,(m/s)$^2$.

\paragraph{Motion-family sensitivity.}
We separately summarize the two fixed 128-world families after
confirmation. In layout order $+-/++/--$, RMSE is
$.22254/.33954/.35324$\,m/s for random-direction motion and
$.23088/.37216/.35454$\,m/s for near-hard motion. Inverting the same
e-value with $\delta=.05/4$ gives upper endpoints
$+.00213/-.00686$\,(m/s)$^2$ for random-direction contrasts and
$-.01528/-.00439$\,(m/s)$^2$ for near-hard contrasts, with at least
95\% simultaneous coverage across these four post-hoc comparisons.
Let $w$ be the near-hard weight, keeping both conditional
laws and the estimator fixed. The simultaneous family
endpoints $U_N,U_G$ give $wU_N+(1-w)U_G$ as an upper
bound on each mixture contrast for all $w\in[0,1]$ on the
same coverage event. Both repeat contrasts have negative
upper bounds for $w>.123$.

For the common $.4$\,m/s projection, a separate $95\%$ family
uses $b=.65$ and $\delta=.05/6$. All six endpoints are below
$-.00220$\,(m/s)$^2$, certifying lower projected mean loss
than both projected repeats and zero at every $w$.

\paragraph{Paired win probabilities.}
For each main Table~2 row and each of $++$, $--$, and zero, we count
$L_{+-}<L_R$ across the shared-noise pairs, with ties as nonwins.
This is a post-hoc family of 18 contrasts. Independent worlds
give one-sided Hoeffding lower bounds $w/256-\sqrt{\log(18/.05)/512}$
for a win count $w$, jointly at $95\%$ by Bonferroni. For the
mixture the target is the equal-family probability. Counts range
from 181 to 219, giving all lower bounds above $.599$. The source
package contains every count, including $195/256$ raw $+-$
wins over zero in the mixture.

\nobalance
\section{Four-view capture}
\label{app:M}
\setcounter{table}{0}\setcounter{figure}{0}\renewcommand{\thefigure}{M.\arabic{figure}}
\renewcommand{\thetable}{M.\arabic{table}}

\begin{figure*}[t]
\centering
\includegraphics[width=.88\textwidth]{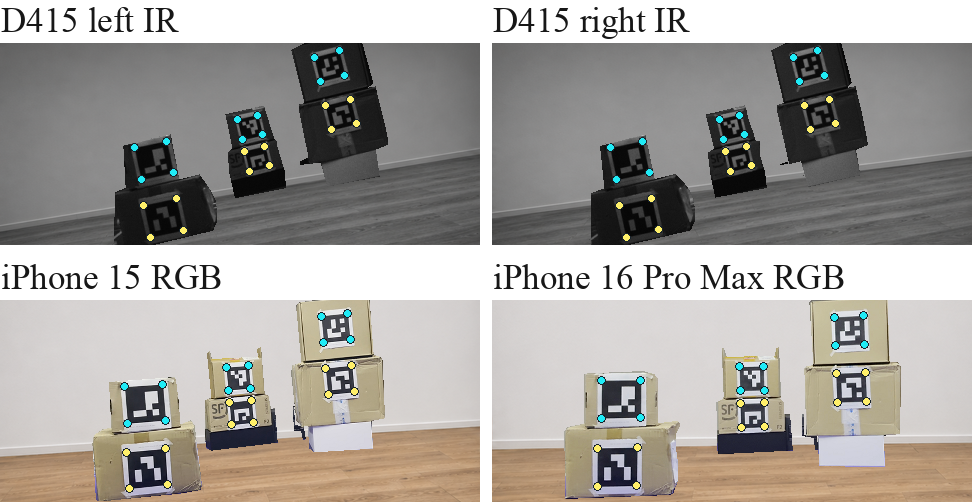}
\caption{Four-view coverage in take A: D415 infrared images and two
iPhone RGB images. AI-generated wood-floor and white-wall backgrounds
are for display; they are not used in measurement or fitting.
Cyan marks velocity-fit corners; yellow marks D415 stereo reference
corners.}
\label{fig:original-capture}
\end{figure*}
\paragraph{Coverage and timing.}
Two handheld takes use a D415 stereo IR pair and two iPhones observing
six fixed fiducials. Marker motion aligns phone presentation times to
D415 sensor times; physical scan directions remain unverified.
Each take has 210 unique four-view sets sampled at 6\,Hz over
10--45\,s: all have at least twelve common corners and 174 have 24.
All 3,598 stereo IR pairs agree on device timestamps and exposure width
(31.979-ms integration, 33.321-ms median cadence). Native line delay
and row mapping are unmeasured.

\paragraph{Disjoint timing transfer.}
In the iPhone 16 Pro Max 4K/120-fps calibration clip, we fit intrinsics
on 57 frames per training partition and signed span in its 19 blocks.
The other 19 blocks are held out, using disjoint corner subsets for
pose fitting and prediction at $1920\times1080$.
\begin{table}[H]
\centering\small
\caption{Disjoint-frame phone timing. Cubic trajectories.
Span [ms] uses decoded coordinates. Scores are RMS [px].}
\label{tab:disjoint-phone}
\setlength{\tabcolsep}{4pt}
\begin{tabular}{@{}llrrrr@{}}
\toprule
Split & Axis & Span & Fitted & Zero & Negated \\
\midrule
Forward & $x$ & $+1.90$ & .16339 & .16345 & .16364 \\
Forward & $y$ & $-.70$ & .16312 & .16345 & .16382 \\
Reverse & $x$ & $+1.02$ & .23067 & .23078 & .23095 \\
Reverse & $y$ & $-2.81$ & .22982 & .23078 & .23327 \\
\bottomrule
\end{tabular}
\end{table}
Vertical cubic fits improve 13/19 and 10/19 blocks over zero. Both gains
over negated span survive single-block deletion; the reverse split's
gain over zero does not. Transferring spans $.70/2.81$\,ms to 31 capture
blocks per take changes RMS versus zero by $+.00025/+.00105$\,px in A
and $-.00012/+.00008$\,px in B at $960\times540$. The $.70$-ms effect
has opposite signs in A and B; quadratic fits agree.

\paragraph{Stereo-referenced velocity.}
We fit twelve corners from three tags and form a velocity reference
from the other three. A local quadratic differentiates D415 camera
centers while omitting each scoring frame and its two neighbors on
either side. Motion-derived phone offsets exclude samples within
$.45$\,s of scoring epochs. Fit and reference share stereo calibration
and timestamps. Disjoint tag references differ by $.010/.006$\,m/s RMS
in A/B; the across-take speed difference is $.044$\,m/s.
\begin{table}[H]
\centering\small
\caption{Velocity RMSE (m/s) relative to the D415 stereo reference at
five elapsed times per take: 18, 21, 24, 27 and 33\,s. We select the
lowest-residual solver-terminated fit. The fit and reference share
stereo calibration and timestamps.}
\begin{tabular}{@{}lrr@{}}\toprule
Condition & A & B\\\midrule
Zero span, free offsets and mounts & .150 & .135\\
Zero-velocity predictor & .054 & .023\\\bottomrule
\end{tabular}
\end{table}
The zero predictor is best in both takes. These recordings establish
four-view coverage and timing-transfer measurements, rather than
complementary-layout velocity gains.

\paragraph{Scan support.}
For projected span $S$, minimum larger-band width $B$ over two-band covers and centered
radius $R$, the minimum of $(B-R/50)_+/(S+R/50)$ over 180 axes bounds
$B/S$ below, since both widths are $2R$-Lipschitz in direction.
Shared corners give $>.24$ allowing 1-px errors; fixed fit corners give
$>.29$ in every eligible set (182 in A, 180 in B). The Y/X aliases give
$\leq.053$ and $5/11$ on their scan axes.

\nobalance
\section{Known anchor timing offset}
\label{app:N}
\paragraph{Static rank lift.}
We retain the anchor rotations and baseline and give A1 a supplied
offset $e$ in normalized time, while A0 retains offset zero.
The anchor chart continues to use A0's exposure. For point $Y_i$,
put $x_i=Y_{i1}/Y_{i3}$ and
$D_{1i}=D\pi(Y_i-\beta e_2)$. In the static augmented matrix
$A(e)$, the change is confined to the A1 motion columns. In order
$(\omega,\bar v,\bar a,B)$ the four blocks of the change are
\[
\begin{split}
 &-2eD_{1i}[Y_i]_\times,\qquad -eD_{1i},\\
 &-\tfrac12\kappa_i(e)D_{1i},\qquad
 -\kappa_i(e)D_{1i}[Y_i]_\times,\\
 &\kappa_i(e)=(x_i+e)^2-x_i^2.
\end{split}
\]
Indeed the static row factor is one, and differentiation at A1
subtracts the trajectory terms at the A0 exposure $x_i$ from
those at $x_i+e$. The scene, mount and jet blocks are unchanged.

For the rational fixture of Appendix A at $y=0$ and $y=1/20$,
we verify $\operatorname{rank} A(0)=71$ for each of $++,+-,--$
using the exact null $n$ and a modular minor. Appending the column
$A'(0)n$ raises the rank to 72 in every case. The certificate uses
the prime 1000003. At each of $e=\pm1/100$, direct modular ranks
are also 72. Run \texttt{static/verify\_anchor\_skew.py} to reconstruct
the matrices and checks.

\paragraph{Consequence for the local rate.}
A basis with last vector $n$ makes the last column of $A(0)$ zero.
The other 71 columns together with $A'(0)n$ are independent,
so some maximal minor in this basis has a nonzero linear term in $e$.
Thus $A(e)$ has full column rank for all sufficiently small
nonzero $e$. For each fixed such $e$, the static finite-clock shear
remains valid. Write $\Delta A=A(e)-A(0)$, which reads only motion components.
Then $\Delta A(U_\tau-I)=0$ and $\Delta A V_s=0$ because both shear
corrections and clock columns have zero motion components. The
cross-clock identities and remainder construction of K.6 therefore give a positive target secant on a
sufficiently small local physical class. Class-global least squares
gives $O(\sigma^2)$ normalized velocity risk for all three layouts.
The constants and class radius here may depend on the supplied
offset. The calculation addresses a known timing intervention at
these fixtures. It does not bound performance with unknown anchor
timing and gives no ordering of the risk constants as $e$ tends to zero.

\paragraph{Finite effect on the fixed witness.}
We hold the $++$ corrected-path endpoint pair of main Section~7.1 fixed
($y=.2$\,m, a 4-ms auxiliary clock step and a $.05$\,m/s
velocity gap) and change only the supplied A1 offset in both
forward means. At $.1$-pixel Gaussian noise the balanced optimum
decision error falls from .406 at simultaneous anchor exposure
to .341, .046 and .00040 at offsets of .1, .5 and 1\,ms.
The means are evaluated with 80-digit implicit roots.
The complementary pair stays separated in all four cells
(noise-normalized distances 44.75--45.25). These values isolate the
sensitivity of this particular ambiguity pair. At a given offset
other physical pairs can remain harder to distinguish.
The calculation uses \texttt{static/anchor\_skew\_pairs.py}.

\paragraph{Noise--offset crossover for the repeated pair family.}
Let $D(\alpha,t,e)$ be the whitened mean difference on the
corrected repeated path of main Section 4.1, with the
same supplied A1 offset $e$ at both endpoints.
On a common compact safe branch with all exposure times in $I$, $D$ is smooth in all three variables.
For each $e$, it vanishes at $\alpha=0$ and at $t=0$. Hence
$\partial_eD$ also vanishes on those two faces.
Integrating the bounded derivative
$\partial_\alpha\partial_t\partial_eD$ gives
\[
 \|D(\alpha,t,e)-D(\alpha,t,0)\|
 \leq C_1|\alpha te|.
\]
The synchronous cancellation gives $\|D(\alpha,t,0)\|\leq
C_0\alpha^2|t|$. For a fixed admissible $t=\rho_{\mathrm{eff}}>0$
and fixed nonzero $y$, take $\alpha$ proportional to
$\min\{\alpha_{\max},\sqrt{\sigma/t},\sigma/(t|e|)\}$, with the last
entry omitted at $e=0$. Here $\alpha_{\max}$ bounds the included
path segment. The two Gaussian means then have bounded KL, while
the metric velocity gap is $2|y|(L/T)\alpha t$. Thus, on any class
containing this path segment, the repeated-layout squared risk is
bounded below up to a positive geometry constant by
\[
 (L/T)^2y^2\min\{t^2\alpha_{\max}^2,\ \sigma t,\ \sigma^2/e^2\}.
\]
This is a lower bound from the specified pair family. At fixed nonzero $e$ the lower envelope is eventually quadratic,
consistent with the local full-rank result on suitable classes.
With physical offset $E=Te$, the crossover occurs at a pixel-noise
scale of order $ft(E/T)^2$ when the path amplitude cap is inactive.
For an unknown offset interval the bound holds for any common
admissible $e$ whose two endpoints lie in the class.

\nobalance
\section{Finite ambiguity under anchor timing offsets}
\label{app:O}
\renewcommand{\thetable}{O.\arabic{table}}
\setcounter{table}{0}
\paragraph{Matched pair search.}
We use the 12 points and corrected $++$ path of Appendix N as a
common starting fixture for $++$ and $+-$. The axial velocity
gaps are $.01,.025,.05,.10$\,m/s. At each gap and layout we hold
the reference endpoint's other physical parameters fixed. We keep the
alternative velocity fixed and refit its other 59 physical coordinates.
Each endpoint's A1 timing is a parameter within the same interval.
A width of zero supplies the timing exactly. For positive widths,
the two hypotheses can use different A1 times. A0 remains at time
offset zero in all conditions.

We use component radii of $2$\,cm for point and mounting coordinates,
$.02$\,rad for auxiliary pose vectors, and $2$\,ms for auxiliary clock
corrections, centered at the alternative path state. Translational and
angular acceleration coordinates have normalized radii $.02$ and
$.03$. The bounded angular-rate parameterization enforces the
$D=.15$ cone. All trials use the same three deterministic starts
and budget of 150 evaluations. We select the smallest distance among
feasible starts and returns and replay each selected pair with
independent 80-digit implicit projection. All 64 search cells and 192
starts are retained. Reprojection coordinates agree within
$1.2\times10^{-16}$ in normalized units.

\begin{table}[H]
\centering\small
\caption{Returned image distances in pixels at a $.10$\,m/s velocity
gap. The center and half-width specify each endpoint's A1 interval
in ms. These are distances of feasible pairs.}
\begin{tabular}{@{}rrrr@{}}
\toprule
Center & Half-width & $++$ & $+-$ \\
\midrule
0 & 0 & .000143 & .7941 \\
0 & .1 & .000143 & .7588 \\
.5 & 0 & .005468 & .8174 \\
.5 & .1 & .003833 & .7790 \\
.5 & .5 & .000143 & .7602 \\
1 & 0 & .080482 & .8697 \\
1 & .1 & .056312 & .8233 \\
1 & .5 & .005321 & .7859 \\
\bottomrule
\end{tabular}
\end{table}

\paragraph{Finite ambiguity after profiling.}
At $.1$-pixel noise, all eight $++$ timing conditions return a
$.10$\,m/s pair within one noise unit, with optimal balanced testing
error at least .343. The two-point inequality of L.5 therefore
gives a worst-case root-risk lower bound of $.0414$\,m/s in each
$++$ condition on classes containing the corresponding pair.
For $+-$ the largest tested gap found within one noise unit is
$.01$\,m/s in all eight conditions. At $.01$\,px, $++$ reaches
$.05$ or $.10$\,m/s and no $+-$ trial passes the threshold.
At $1$\,px both layouts reach the largest tested gap. These searches provide feasible witnesses
and upper bounds on the minimum image distances on the searched
slices. They give no lower bound on those distances or comparison
of full-class risks.

In particular, the fixed $.05$\,m/s $++$ pair in Appendix N
becomes easier to distinguish at $.5$\,ms. At the same supplied
timing and velocity gap, refitting the alternative lowers its
distance from $.338$ to $.00182$\,px. The local full-rank result
of Appendix N still applies on its stated smaller classes.
The timing intervals here are sensitivity settings for a simulated
fixture and have no calibration interpretation for the handheld
captures.

Sharing one free A1 offset between hypotheses preserves the largest
attained gaps in all 40 nonzero-width cells. At $.10$\,m/s and interval
$[.5,1.5]$\,ms, $++$ has distance $.00547$\,px with shared offset
$.5$\,ms, versus $.00532$\,px with separate offsets.

\paragraph{A broad-class bound with supplied offsets.}
Use Appendix X's $\pm.4$\,m/s pair with a common supplied clock
tuple across the two states. From their nominal offsets, shift A1 by
at most $1$\,ms and each auxiliary independently by at most $.25$\,ms.
This near-hard pair has axial acceleration $106.7$\,m/s$^2$.
For $|t|\leq.58$, depths exceed $4.29$\,m and numerators
are at most $1.2$ in magnitude. Both projection derivatives
satisfy $|\partial_t\pi_j|\leq\beta=.04$, making the row map a
contraction into the full raster.
For normalized clock shift $\eta_c$, both roots move from the
original exposure $t_0\in\{r_i,s_i\}$ by at most
$g=|\eta_c|/(1-\beta)$. The new image rows move by at most $\beta g$,
so gaps exceed $15.61$\,px and spans exceed $12.016$\,ms.
At a common time within $g$ of $t_0$, half the depth difference
is at most $h=mg(g+|r_i-s_i|)$. With shared numerator rate
at most $.0655$, the same-time coordinate gap is at most
$G_j=2h(|N_j(t_0)|+.0655g)/(z_i-h)^2$.
The two root observations differ by at most $G_1/(1-\beta)$
and $G_2+\beta G_1/(1-\beta)$ in their respective coordinates.
The rational checker \texttt{check\_uniform\_row\_skew.py} sums
their squares and scales by $(768/.1)^2$ for independent
$.1$\,px Gaussian noise. The resulting distance is $d<1$ for all
layouts ($d<.548$ for A1 skew up to $.5$\,ms).
Under the equal prior, the posterior variance is $.4^2\operatorname{sech}^2(dX/2)$.
The mixture density is $\phi(X)e^{-d^2/8}\cosh(dX/2)$ for standard
normal density $\phi$. Integrating with $\operatorname{sech} u\geq e^{-u^2/2}$
gives squared Bayes risk at least $.4^2e^{-d^2/8}/\sqrt{1+d^2/4}$.
Thus minimax root risk exceeds $.355$\,m/s ($.385$\,m/s at $.5$\,ms).

\nobalance
\section{Anchor timing and translation observability}
\label{app:P}
\renewcommand{\thetable}{P.\arabic{table}}
\setcounter{table}{0}

\paragraph{Paired protocol.}
We draw 128 twelve-point worlds from Appendix L.12's
laws, with 32 per motion family and A1 offset ($.5$ or
$1$\,ms). A0 remains at zero. Layouts and synchronized
controls share worlds and noise. The four equal-weight
strata are conditioned on a joint positive-depth, simple
row and observation-dependent fitting-box gate across
layouts and offsets. All worlds pass without replacement.
Pre-gate Gaussian noise has $.1$\,px standard deviation.

We supply anchor calibration and scan laws. At nonzero
offset, we fit 62 coordinates with timing supplied or 63
with A1 in $[-1.5,1.5]$\,ms, starting that offset at zero.
Both arms initialize from anchors and select by residual
over two starts (300 evaluations each). A zero fallback
covers cases without a finite fit. We also report projection
onto the $.4$\,m/s speed ball.

Four primary comparisons measure paired raw excess loss
for $+-$ versus each repeat under each timing treatment.
We invert the L.12 e-value ($b=1.45$, $\delta=.05/4$) for
simultaneous $95\%$ upper endpoints. The synchronized
control, projected loss and across-treatment changes are
descriptive.

\begin{table}[H]
\centering\small
\caption{Velocity RMSE in m/s on the same 128 worlds. Parentheses
give results after projection onto the common speed ball.}
\begin{tabular}{@{}lrrr@{}}
\toprule
A1 timing & $++$ & $--$ & $+-$ \\
\midrule
Synchronized & .362 (.301) & .379 (.300) & .220 (.203) \\
Nonzero, supplied & .351 (.293) & .345 (.278) & .219 (.203) \\
Nonzero, estimated & .356 (.296) & .353 (.286) & .218 (.202) \\
\bottomrule
\end{tabular}
\end{table}

\paragraph{Timing upper bound with unknown auxiliaries.}
At synchronized anchors, let $G_{ci}=D_{P_i}\mu_{ci}$
include implicit-row feedback. A velocity step $w$ and
acceleration step $h$ can preserve the anchor images to
first order via $\delta P_i=r_i w+r_i^2h/2$. At exact
exposure $t_{ci}$ the displacement step is
$-\Delta_{ci}(w+q_{ci}h)$, where $\Delta_{ci}=t_{ci}-r_i$ and
$q_{ci}=(t_{ci}+r_i)/2$. The image step is $G_{ci}$ times
that displacement. Stack $\Delta_{ci}G_{ci}$ and
$\Delta_{ci}q_{ci}G_{ci}$, whitening both stacks by
$\Sigma^{-1/2}$ to get $U,W$. Minimizing over $h$ in
this subset of nuisance steps gives
\[
\sigma^2 I_v\preceq U^T(I-\Pi_W)U.
\]
Here $\Pi_W$ projects onto the column space of $W$, and
$I_v$ retains all 59 nuisances. A common nonzero $q$ on
rows with $\Delta\ne0$ gives $W=qU$ and a zero bound.
The bound uses the state's latent times and Jacobians.
Across 256 worlds per layout, the median of
$\lambda_{\min}(\sigma^2I_v)/\lambda_{\min}(U^T(I-\Pi_W)U)$
is $5.9/5.6/13.1\%$ for $++/--/+-$. We replay exact means
with $\Sigma=I$ (Appendix L.12).

\paragraph{Velocity estimates with nonzero anchor offsets.}
With the offset supplied, simultaneous upper endpoints on the
complementary-minus-repeated loss are $-.00764$ and
$-.00218$\,(m/s)$^2$ for $++$ and $--$. With the offset
estimated, the corresponding endpoints are $-.01079$ and
$-.00774$\,(m/s)$^2$. These bounds jointly support lower expected
raw squared loss for the complementary layout under both
timing treatments in the tested population. The zero predictor
has RMSE $.295$\,m/s. All 1,152 conditions select a finite state
and meet the $10^{-5}$ column-scaled projected-gradient threshold.

\paragraph{Exact rank test with supplied calibration.}
With exact pixels, supplied $R(t),Q_s,b_s,\tau_s$ and synchronized
anchors, metric disparity gives $Y_i$ at anchor time $r_i$.
Use the known-rotation bearing incidence of Su et al.~[18].
For auxiliary time $t$, let $d$ be its unit world bearing and
$H=I-dd^T$. Substituting $P_i=C(r_i)+R(r_i)^TY_i$ gives
\[
\Delta H(v+q a)=H(R(r_i)^TY_i-R(t)^Tb_s),
\]
where $\Delta=t-r_i$ and $q=(t+r_i)/2$. Stack the velocity
and acceleration blocks as $A$ and $D$, and project out
$\operatorname{im}D$. Velocity is unique on the safe branch
when $\operatorname{rank}((I-\Pi_D)A)=3$. At an interior
state this is necessary: a rank defect supplies a nonzero
velocity change and a compensating acceleration change.
Small steps preserve depth and the selected simple row roots.

Two distinct midpoint groups, each with two nonparallel
world bearings at $\Delta\ne0$, fix both $v$ and $a$.
A single such group at $q=0$ fixes $v$. For any $q_*\ne0$,
the trial acceleration increment $-w/q_*$ gives
\[
\|(I-\Pi_D)Aw\|
\leq \frac{\big[\sum_j\Delta_j^2(q_j-q_*)^2\|H_jw\|^2\big]^{1/2}}
{|q_*|}.
\]
All active rows in Appendix X have $q_j=k/2$, giving rank
zero. At the exact means of all 768 Appendix L.12
layout--world pairs, numerical rank is three and recovered
velocity errors are below $3\times10^{-12}$\,m/s. This is a fixed-data
geometric test with supplied calibration.

\nobalance
\section{Public rolling-shutter data diagnostics}
\label{app:Q}
\renewcommand{\thetable}{Q.\arabic{table}}
\setcounter{table}{0}

\paragraph{Motion coverage from external tracks.}
We audit the ten published OptiTrack trajectories of the TUM
Rolling-Shutter VIO benchmark [25].
We estimate derivatives at 20\,Hz using temporally supported
local cubic fits to the position of the IMU origin in the
motion-capture frame and
to center-relative rotation vectors. At each time, we compute the smallest cone
constant consistent with the estimated motion,
\[
 D_{\rm req}=\frac{\|\Omega_0\|}
 {T\sqrt{4\|A_{\rm tr,0}\|^2/L^2+\|\Gamma_0\|^2}}.
\]
We use $T=.05$\,s and $L=1$\,m, matching the simulation units.
This checks the motion condition used in simulations.
The theoretical local class permits other fixed finite values of
$D$, with the certified class depending on the choice.

\begin{table}[H]
\centering\small
\caption{Motion coverage across the ten TUM trajectories.
The window is the half-width of the local derivative fit.
Counts refer to sampled times.}
\begin{tabular}{@{}lrrrr@{}}
\toprule
Window & Times & $D_{\rm req}\leq .15$ & $D_{\rm req}\leq1$ & Median $D_{\rm req}$ \\
\midrule
50\,ms & 9458 & 2 & 491 & 2.86 \\
75\,ms & 9459 & 2 & 363 & 3.16 \\
100\,ms & 9459 & 2 & 282 & 3.42 \\
\bottomrule
\end{tabular}
\end{table}

In the primary 75-ms window, 484 times have speeds between
$.16$ and $.4$\,m/s. None meet the $.15$ cone at the original
sampling phase, and one does at each of three shifted phases.
These indoor trajectories seldom visit that specific
low-initial-angular-rate regime at the evaluated cadence.
The results use differentiated, aligned motion-capture poses
and depend on derivative bandwidth. They describe motion coverage
under the chosen thresholds, with geometric, clock and other class
membership conditions outside this check.

\nobalance
\section{Simulated layout comparison on tracked motion}
\label{app:R}
\renewcommand{\thetable}{R.\arabic{table}}
\setcounter{table}{0}

\paragraph{Paired trajectory replay.}
We select six windows uniformly by eligibility rank from each of ten
TUM OptiTrack sequences: 60 windows among Q's 484 eligible times.
Local cubic body-frame translation and world-to-body Cayley rotation
supply motion; cone ratios are $.22$--$9.92$ (median $2.20$).
We synthesize twelve-point observations with $L=1$\,m, $T=.05$\,s,
$.1$-px noise, point depths 2--5\,m, true clocks $\pm4$\,ms and
$\pm2$-mm mount jitter, paired across layouts.

We fit all 62 unknowns with two nominal-mount starts and 300 evaluations
per start, selecting by image residual. Bounds are
$|\bar v_j|\leq.04/\sqrt3$, $|\bar a_j|,|B_j|\leq.12$,
$|\omega_j|\leq.25$ and clocks $\pm6$\,ms. All selected returns
are stationary with valid in-raster roots. Median cubic-to-quadratic
image mismatch is $.0029$--$.0034$\,px RMS. Aggregate RMSE is
$.343/.367/.199$\,m/s ($++/--/+-$), versus $.293$ for zero.
Final per-sequence rows are in \texttt{data/tum\_replay\_per\_sequence.csv}.
\paragraph{Calibration sensitivity.}
Shifting all A1 ordinates at fixed exposure times by
$-1,-.25,+.25,+1$\,px gives complementary RMSE
$.200/.199/.198/.198$\,m/s. Across these four biases, the worst mean-loss
gaps after any three sequence deletions are
$-.043/-.063/-.039$\,(m/s)$^2$ against $++/--/$zero.
With one $1.5$-px error in each camera, the worst sequence-deletion
$L^1$ gaps remain negative. These controls retain fixed exposure times.

\subsection{Replays with triangulated scene points}
We replace random points by the first twelve points visible in all
three layouts in a fixed seeded order, triangulated in the metric GS
frame at each center epoch. The 37 eligible windows cover nine TUM
sequences; point depths remain 2--5\,m. Motion, mounts, clocks and
noise draws are unchanged, with zero depth shift and $\pm6$-ms fitted
clocks. Table~\ref{tab:real-point-replay} pairs the two point sets.
\begin{table}[H]
\centering\small
\caption{Velocity RMSE (m/s) on the same 37 windows. Each
column uses the same motions and noise draws for both point sets.}
\label{tab:tum-measured-points}
\label{tab:real-point-replay}
\begin{tabular}{@{}lrrrr@{}}
\toprule
Scene points & $++$ & $--$ & $+-$ & Zero \\
\midrule
Random       & .333 & .364 & .181 & .302 \\
Triangulated & .533 & .628 & .497 & .302 \\
\bottomrule
\end{tabular}
\end{table}
Median narrowest-principal-axis scene spread drops from $.313$ to
$.046$\,m. Complementary local linearized velocity-noise RMS rises
from $.199$ to $.772$\,m/s, while typical image residual changes from
$.059$ to $.061$\,px. Coordinate-noise loss therefore depends strongly
on the observed geometry; zero prediction gives $.302$\,m/s on this set.

\nobalance
\section{Scan-slope sensitivity on a public GS/RS rig}
\label{app:S}
\renewcommand{\thetable}{S.\arabic{table}}
\setcounter{table}{0}

We vary timing on fixed pixels. Orthogonal reorientation
$F$ maps pixels, scan covector and covariance $(u,\ell,\Sigma)$
to $(Fu,F\ell,F\Sigma F^\top)$. This preserves
$\tau+\ell^\top u$ and whitened squared errors.

\paragraph{Fixed-image protocol.}
We evaluate 108 frame groups: 18 in sequence 4 and ten in each other
TUM sequence. Outer GS images at $\pm150$\,ms and OptiTrack poses
triangulate mutual SIFT tracks, with depth $.3$--$30$\,m, GS error below
$2$\,px and parallax above $.75^\circ$. Center GS PnP requires twelve
tracks. Seven failed gates leave 32,135 tracks in 101 groups.

We fit constant world-frame IMU-origin velocity and body-frame angular
rate, bounded componentwise by $\pm3$\,m/s and $\pm12$\,rad/s, using
implicit distorted rows, soft-$L^1$ loss ($2$\,px), zero initialization
and 200 evaluations. The $29.4737$-$\mu$s line delay gives a
$15.0758$-ms center-row offset. Nominal, zero and negated slopes keep
that center time fixed. We score against the derivative of a tracked
position cubic in the central $\pm75$\,ms, disjoint from triangulation
poses. Equal-sequence RMSE changes by at most $.002$\,m/s under
$\pm50/100$-ms scoring windows.
\begin{table}[H]
\centering\small
\setlength{\tabcolsep}{3pt}
\caption{Velocity RMSE in m/s against OptiTrack. The three
timing fits share the sensor center-row time. Doc. uses the published slope. Secant uses the
supplied outer OptiTrack positions. The last two rows weight
sequences equally.}
\begin{tabular}{@{}lrrrrr@{}}
\toprule
Groups & Doc. & Zero slope & Reversed & Zero vel. & Secant\\
\midrule
Seq.~1 ($9$) & .38 & 3.65 & 4.03 & 1.15 & .082 \\
Seq.~2 ($9$) & .37 & 3.07 & 3.87 & 1.44 & .070 \\
Seq.~3 ($10$) & .60 & 2.18 & 3.02 & 1.06 & .047 \\
Seq.~4 ($18$) & 1.25 & 2.29 & 3.33 & .87 & .037 \\
Seq.~5 ($9$) & .61 & 1.32 & 2.80 & .76 & .027 \\
Seq.~6 ($10$) & .40 & 3.04 & 3.86 & 1.08 & .043 \\
Seq.~7 ($9$) & .44 & 3.01 & 3.91 & 1.00 & .038 \\
Seq.~8 ($9$) & .25 & 3.33 & 4.01 & .93 & .034 \\
Seq.~9 ($8$) & .54 & 3.33 & 4.08 & .84 & .062 \\
Seq.~10 ($10$) & 1.79 & 3.47 & 4.14 & .83 & .083 \\
\midrule
Pooled ($101$) & .87 & 2.89 & 3.69 & 1.00 & .055 \\
All 10 & .80 & 2.95 & 3.73 & 1.01 & .056 \\
New 7 & .80 & 2.96 & 3.71 & .98 & .055 \\
\bottomrule
\end{tabular}
\end{table}
All slope fits terminate with valid in-raster rows. The nominal slope
beats both slope controls in every sequence and zero velocity in
79/101 groups and eight sequences. The largest nominal-slope group
errors are $4.96$ and $3.35$\,m/s in sequences 10 and 4.

\paragraph{Supplied-pose reference and held-out fusion.}
The outer OptiTrack positions also supply a $300$-ms secant velocity,
with endpoints outside the central reference window of the same tracker.
It beats the nominal-slope fit in 100/101 groups and every sequence.
Over all 108 attempts, slope fits with zero fallback give
$.857/2.851/3.604$\,m/s (nominal/zero/reversed).
For $v_{\rm sec}+\alpha(v_{\rm RS}-v_{\rm sec})$, each held-out sequence
uses the $\alpha\in[0,1]$ minimizing equal-sequence MSE on the other nine;
image-gate failures use the secant. RMSE is $.058$\,m/s versus $.056$
for the secant. Nine folds choose $\alpha=0$; sequence 10 chooses
$.0123$, changing its error from $.083$ to $.095$\,m/s. The comparison
measures the incremental RS fit given the supplied tracker inputs.

\nobalance
\section{Anchor spread and paired scene contraction}
\label{app:T}
\renewcommand{\thetable}{T.\arabic{table}}
\setcounter{table}{0}

\paragraph{Observed spread and paired contraction.}
Calibrated synchronous anchors initialize
$\widehat Y_i=\beta(u_{0i},v_{0i},1)/(v_{0i}-v_{1i})$.
We measure spread by the least singular value of their centered
$12\times3$ point matrix divided by $\sqrt{12}$, without auxiliary
observations or fitted motion.

We generate 64 scenes, 32 per L.12 motion family, and contract point
offsets about their centroid by $1,.25,.05$. Motion, mounts, clocks,
point identities and Gaussian noise are paired across factors/layouts.
Noncoplanarity persists. The same 96-coordinate, 62-unknown estimator
uses two starts and 300 evaluations per start; all returns are scored.
\begin{table}[H]
\centering\small
\setlength{\tabcolsep}{4pt}
\caption{Velocity RMSE in m/s across the 64 new paired
scenes at $.1$\,px. Spread is the median anchor-derived
score in m.}
\begin{tabular}{@{}lrrrrr@{}}
\toprule
Factor & Spread & $++$ & $--$ & $+-$ & Zero \\
\midrule
$1$   & .347 & .327 & .345 & .220 & .278 \\
$.25$ & .086 & .727 & .775 & .761 & .278 \\
$.05$ & .017 & .805 & .865 & .841 & .278 \\
\bottomrule
\end{tabular}
\end{table}
From factor $1$ to $.05$, complementary-minus-repeated mean loss
increases by $.118/.030$\,(m/s)$^2$, with nominal 97.5\% stratified
world-bootstrap intervals $[.016,.219]/[-.061,.118]$ (20,000 draws).
At $.05$, selected fits reach the budget in $53/54/55$ of 64 cases,
exceed the stationarity threshold in $55/61/61$, and hit the velocity
box in $64/61/62$ ($++/--/+-$). Common $.4$-m/s speed projection gives
$.461/.520/.503$\,m/s, still above zero's $.278$. These scores describe
the fixed estimator and budget under a geometry intervention.

\nobalance
\section{Accuracy of supplied auxiliary timing}
\label{app:U}
\renewcommand{\thetable}{U.\arabic{table}}
\setcounter{table}{0}

\paragraph{Which clock report restores quadratic risk?}
For repeated \(++\), let a unit vector \(\ell\in\mathbb R^2\)
specify the report \(s\), with \(|s-\ell^\top\tau|\leq\delta_\tau\).
The estimator receives the whitened images \(Y\) and report \(s\).
Risk takes the supremum over physical states and admissible reports.
Clocks use units of \(T\) relative to the anchor time reference.
Put \(\lambda=|\ell^\top(1,1)|\). If \(\lambda>0\),
on a sufficiently small fixed local product and for small \(\sigma\),
\[
\begin{aligned}
 R_\ell^*&\lesssim (L/T)^2(\sigma^2+\delta_\tau^2),\\
 R_\ell^*&\gtrsim (L/T)^2\max\{\sigma^2,
            \min(\bar\alpha^2 t^2,\sigma t)\},\\
 t&=\min\{2\delta_\tau/\lambda,t_0\}.
\end{aligned}
\]
For any \(\kappa>0\), one product and constants suffice
for all \(\lambda\ge\kappa\). Here \(\bar\alpha,t_0>0\)
are fixed admissible path caps. Thus quadratic risk holds
exactly when \(\delta_\tau=O(\sigma)\). If \(\lambda=0\),
the \(\sigma t_0\) lower bound persists even for an exact report.

\paragraph{Uniform upper bound.}
In Appendix K.6's secant write \(D_\ell=\ell_1 J_1+\ell_2 J_2\)
for the weighted jet selector. Then
\(\|D_\ell z\|=\alpha|\ell^\top\Delta\tau|\).
The unit transported null \(n_\tau\) has two identical nonzero
jets, so \(\|D_\ell n_\tau\|\) is bounded below for \(\lambda\ge\kappa\).
Decompose \(z=w+h n_\tau\) with \(w\perp n_\tau\).
The nonzero singular-value bound controls \(\|w\|\) by
\(\|AU_\tau z\|\), and
\(|h|\le(\|D_\ell z\|+\|w\|)/\|D_\ell n_\tau\|\).
Since \(\alpha\) is bounded, absorbing the K.6 remainder
gives \(\|\Delta\bar v\|\lesssim_\kappa
\|\Delta G\|+|\ell^\top\Delta\tau|\).
A measurable \(\sigma\)-approximate minimizer of
\(\|Y-G\|+|s-\ell^\top\tau|\) on a countable dense set
then gives the upper risk bound.

\paragraph{Lower bound.}
A static-reference velocity testing pair at common clocks gives
the \(\sigma^2\) term. Appendix C's endpoints with clocks
\((0,0)\) and \((t,t)\) share report \(s=t\ell^\top(1,1)/2\).
For \(t>0\), use \(\alpha=c\min(\bar\alpha,\sqrt{\sigma/t})\)
with fixed small \(c>0\). The image gap \(O(\alpha^2t)\)
and velocity gap \(2|y|(L/T)\alpha t\) give the path term.
An exact difference reports zero at both endpoints for \(t=t_0\).

\paragraph{Matched exact reports at $.1$\,px.}
Both 61-variable fits keep the clock caps on the 256
worlds. Raw RMSE ($++/--/+- $) is $.326/.326/.219$ for the
mean and $.346/.347/.219$\,m/s for the half-difference.
After $.4$\,m/s speed projection, L.12 upper endpoints
($\delta=.05/7$, $b=.65$) are positive for mean minus
half-difference and below $-.0135$\,(m/s)$^2$ for
complementary minus repeat. Projected axial report-loss
contrasts are $-.00214/+.00022/+.00090$\,(m/s)$^2$; all intervals
include zero.

\begingroup
\setlength{\intextsep}{6pt}
\paragraph{Original-task clock intervention.}
We fix either or both true offsets on the L.12 inputs.
The free-clock fit and the 61- or 60-variable fits share
the other bounds, loss, starts and budget. Write
$G_R=L_{+-}-L_R$. A positive change in $G_R$ means
the complementary advantage shrinks. We use 20,000
paired within-family bootstrap draws with nominal
Bonferroni-adjusted 95\% intervals (six total-loss,
32 directional gaps and interactions, and nine report
directional contrasts, in separate families).

\begin{table}[H]
\centering\small
\caption{RMSE [m/s] on 256 worlds (zero $.27900$).}
\begin{tabular}{lrrr}
\toprule
Supplied clocks & $++$ & $--$ & $+-$\\
\midrule
None & .35623 & .35389 & .22675\\
U1 & .32499 & .32656 & .21787\\
U2 & .32277 & .32481 & .22014\\
Both & .31101 & .31228 & .21090\\
\bottomrule
\end{tabular}
\end{table}

\begin{table}[H]
\centering\small
\caption{Mean $G_R$ change from fitted clocks in (m/s)$^2$.}
\begin{tabular}{lrrl}
\toprule
Supplied & $R$ & Change & Interval\\
\midrule
U1 & $++$ & .01733 & $[.00012,.03409]$\\
U1 & $--$ & .01465 & $[.00352,.02628]$\\
U2 & $++$ & .01977 & $[.00341,.03561]$\\
U2 & $--$ & .01678 & $[.00467,.02969]$\\
Both & $++$ & .02323 & $[.00517,.04046]$\\
Both & $--$ & .02078 & $[.00760,.03403]$\\
\bottomrule
\end{tabular}
\end{table}

All 2,304 refits have two finite starts. Selected fits are
below budget with projected gradients at most $10^{-5}$.
With both offsets, mean gaps are $-.05225/-.05304$\,(m/s)$^2$.
After $.4$\,m/s speed projection, all nine post-hoc upper
bounds versus $++$, $--$ and zero are below $-.0111$\,(m/s)$^2$
($\delta=.05/9$, $b=.65$, Appendix L.12).

\paragraph{Directional loss.}
We split squared loss into body-axis $e_3$ and transverse $e_1,e_2$
components. Supplying both clocks attenuates the raw layout gap by
$5.52/17.72$ and $3.85/16.93$ for mixture $++/--$, and
$8.63/18.56$ and $3.59/15.58$ for random-direction $++/--$
(axial/transverse sum, in $10^{-3}$\,(m/s)$^2$).
All 16 projected component-gap intervals favor complementary readout
across the two clock protocols and two cohorts. All four intervals for
projected axial attenuation minus half the transverse sum include zero.
Full values are in \texttt{data/clock\_directional\_components.csv}.
\endgroup

\nobalance
\section{Depth and auxiliary timing}
\label{app:V}
\renewcommand{\thetable}{V.\arabic{table}}
\setcounter{table}{0}

\paragraph{Joint lower envelope.}
Let $z$ be the common base depth shift of the corrected
$++$ path in Appendix J.1. Both auxiliary centres remain
unknown and may take their path-dependent increments.
Use the same fixed compact admissible rectangle in
$(\alpha,t,z)$, with amplitude cap $\bar\alpha$ and positive
clock cap $t_0$. At fixed $z$, define $R^*_{++,z}$ as in
Appendix U for either choice of reported auxiliary clock,
with worst-case normalized error $\delta_\tau$.
Set $t=\min\{2\delta_\tau,t_0\}$. For any class containing
the shifted path, at $t>0$ the risk obeys
\[
R^*_{++,z}\gtrsim (L/T)^2 y^2
\min\{\bar\alpha^2t^2,\,\sigma t,\,\sigma^2/z^2\},
\]
with the last term omitted at $z=0$. At $t=0$ this
path contribution is zero. The constant is uniform on the
chosen rectangle.

Indeed, either reported clock can take the common value
$t/2$ at the endpoints $0$ and $t$. Equation (61) bounds
their whitened image gap by $C\alpha t\sqrt{\alpha^2+z^2}$.
Choose $\alpha=c\min\{\bar\alpha,\sqrt{\sigma/t},
\sigma/(t|z|)\}$, omitting the last term at $z=0$.
Then the image gap is bounded by $C\sqrt{c^4+c^2}\,\sigma$.
The endpoint velocity gap remains $2|y|(L/T)\alpha t$,
so the Gaussian two-point bound gives the stated inequality.
Before the amplitude cap binds, the two terms cross at
$\sigma\asymp t z^2$. This crossover describes the
constructed lower envelope.

Along $z(\sigma)\to0$ with fixed nonzero $y$ and uniform
positive path caps, quadratic risk still requires
$\delta_\tau=O(\sigma)$. If $\delta_\tau/\sigma$ is unbounded,
then $t/\sigma$ is unbounded along a subsequence. Each term
in the minimum, divided by $\sigma^2$, diverges on that
subsequence.

\nobalance
\section{Velocity compatibility diameter}
\label{app:W}
\renewcommand{\thetable}{W.\arabic{table}}
\setcounter{table}{0}

\paragraph{Compatibility under a spread constraint.}
Use the fixed physical envelope of Appendix L.10 with
$L=1$\,m, $T=.05$\,s and $f=768$\,px. Require all roots
to be uniquely defined on the sensor raster with positive
depth, and require scene spread $s(P)>.10$\,m, where
$s(P)$ is the least singular value of the centred point
matrix divided by $\sqrt{12}$. Given $Y$, keep states with
$\|f(\mu_d-Y)\|_2\leq .1\sqrt{\chi^2_{96}(.95)}$.
The velocity cube caps the diameter at $1.6$\,m/s and
the axial diameter at $1.6/\sqrt{3}$\,m/s.

\paragraph{An explicit saturated example.}
Take all sign combinations $(\pm.11,.05\pm.11,z)$
at $z=4.8,5.1,5.4$\,m. Their centred covariance is
$\operatorname{diag}(.11^2,.11^2,.06)$, so $s(P)=.11$\,m.
Use nominal mounts, zero clocks and zero angular and
quadratic motion. At both antipodal velocity corners,
$v=\pm(.04/\sqrt3)(1,1,1)$, a camera coordinate at time
$t$ is $(n_j-tw_j)/(z-tv_z)$, where $n_j$ is the static
numerator and $w_j$ the corresponding rotated velocity.
For all layouts, $|n_1|\leq .11$, $|n_2|\leq .26$ and
$|w_j|,|v_z|<u=.0231$. On $[-.5,.5]$ the scan projection
lies in the raster and its derivative has magnitude below
$.006$, since
\[
\left|\partial_t\pi_1\right|
\leq \frac{u(5.4+.11)}{(4.8-.5u)^2}.
\]
Thus the scan equation has one root. It satisfies
$|t|\leq .11/(4.8-1.5u)<a_t=.024$. The deviation of either
projected coordinate from its static value is bounded by
\[
\frac{a_t u(1+.26/4.8)}{4.8-a_t u}.
\]
Multiplying by $768\sqrt{96}$ gives an image displacement
less than $.936$\,px in Euclidean norm. The compatibility
radius exceeds $1$\,px. Indeed,
$\Pr(\chi^2_{96}>100)=e^{-50}\sum_{k=0}^{47}50^k/k!>.05$,
using $e<11/4$ and only the last two terms.

Every observation within $.064$\,px of each layout's
static mean therefore admits both velocity corners.
The exact diameter is $1.6$\,m/s for all layouts.
The static state has this event with positive probability at
$.1$\,px, also when all layouts share the same noise draw.
The noisy anchor triangulation changes by less than $.020$\,m
in Frobenius norm across the twelve points in this ball:
disparity starts at least $1/27$, its Euclidean change is
at most $\sqrt2/12000$, and the lateral ray changes by at most
$1/12000$. Thus its centred RMS spread stays above $.10$\,m.

\paragraph{Interval checks on fixed observations.}
We reuse the 32 twelve-point worlds of Appendix L.10.
For each of eight velocity vertices, we vary the remaining
59 coordinates from two starts: the original anchor start
and the fixed axial witness at the matching sign.
Both use the original projection and a budget of 600
evaluations. The 768 queries retain all 1,536 starts.
We also recheck both starts of the fixed axial endpoints.

At each returned state, use rational comparisons for the
physical limits and $s(P)>.10$\,m. For a candidate near or
below the residual threshold, enclose the signed row derivative
on a cover of $[-.5,.5]$ and bracket the row root with
outward-rounded interval arithmetic. Enclose the second
coordinate and all 96 residuals before testing compatibility.
A verified antipodal pair matches the cube upper bound.
Fits screened above threshold leave the corresponding
vertices unresolved.

\begin{table}[H]
\centering\small
\caption{Compatibility diameters at $.1$\,px. Counts use all 32
worlds per layout. A confirmed full diameter is $1.6$\,m/s.
The axial diameter is $1.6/\sqrt{3}$\,m/s.}
\begin{tabular}{@{}lrrr@{}}
\toprule
 & $++$ & $--$ & $+-$\\
\midrule
Confirmed full diameter & 31 & 30 & 19\\
Confirmed axial diameter & 32 & 32 & 32\\
Interval-verified states & 601 & 580 & 442\\
Large-residual candidates & 39 & 60 & 198\\
\bottomrule
\end{tabular}
\end{table}

All 19 complementary saturations also saturate both repeated
layouts on the paired observations. Of the 25 worlds with
at least one qualifying pair in Figure~3, 15 saturate under $+-$.
The checks include
the point-spread constraint for both endpoints and the same
physical velocity cube in each layout. Remaining cases
have lower witnesses and the common cube upper bound.
The cube includes candidate speeds up to $.8$\,m/s; the
simulation generator uses speeds up to $.4$\,m/s.
The diameter describes a single compatible set.
Appendix L.12 measures the estimator's average risk.

\paragraph{Truth-centered joint witnesses.}
We shift true axial velocity by $\pm.025$\,m/s in the same
32 worlds. Both endpoints pass all three per-layout cutoffs
in all 28 truth-compatible worlds with other coordinates
fixed. Shared noise places the other four truths above the
cutoff. Using the stacked $288\times62$ truth Jacobian,
we also compensate the axial step by least squares over
the other coordinates. Nonlinear replay gives the same 28
passing worlds and full gaps of $.0500$--$.0612$\,m/s.
Both constructions retain 28 pairs at spread above $.10$\,m
and 27 at a $.4$\,m/s speed cap.

We use $\|R-I\|\leq2\|k\|$ and $\|R'\|\leq2\|k'\|$ for
the Cayley vector $k(t)$ to bound full-raster depth above
$1.87$\,m, $|\pi_1|<.310$ and $|\partial_t\pi_1|<.223$.
The row map therefore has a unique in-raster fixed point.
The four misses leave other joint pairs unresolved.

\nobalance
\section{Finite risk with scene spread}
\label{app:X}

\paragraph{Exact ambiguity with uniform rows.}
Use $k=.15$, $\tau=k/2$, $m=2/15$, $B=.05e_3$ and
$\omega=-kB/2$. For $i=0,\ldots,11$, put
$r_i=\tau+(i-11/2)\,.022$, $s_i=k-r_i$ and $j_i=\min(i,11-i)$.
Let $q_i=r_i$ for even $j_i$ and $s_i$ otherwise, and cycle
depths $4.35,4.65,4.95$ by $j_i$. Set
\[
\begin{aligned}
\bar v_\pm&=\pm mk e_3,\quad \bar a_\pm=\mp2m e_3,\\
P_{i,\pm}&=R(r_i)^T(r_i z_i,\,.05-(q_i-\tau)z_i,\,z_i)^T
              \pm m r_i s_i e_3.
\end{aligned}
\]
The displacements differ by $2m(t-r_i)(t-s_i)e_3$.
Since $R(r_i)=R(s_i)$, either state gives the same
rotated displacement at $r_i$ and $s_i$.
With nominal mounts and auxiliary clocks $\tau$, the
anchors expose at $r_i$, positive-readout auxiliaries
at $q_i$, and negative ones at $k-q_i$. These lists permute
the same grid. Each camera has 12 distinct rows, separated by
$16.896$\,px, with span $12.1$\,ms and two-band ratio $5/11$.
Exact minors bound centered spread above $.24$\,m.
The full-raster bounds give depth $>4.29$\,m and
$|\partial_t\pi_1|<.04$ (\texttt{check\_uniform\_row\_alias.py}).

The near-hard law gives both states support under the
joint physical validity gate. Taking $m$ to the speed cap
from below preserves the same means and row grid.
The targets $\pm.4e_3$\,m/s with identical observation laws
give squared minimax risk at least $.16$\,(m/s)$^2$.
Zero prediction gives the upper bound. This holds even on
the uniform-row subclass with both clocks known.

For $m=.07$, the same certificate gives antipodal $.21$\,m/s
targets inside the random-direction support, including its
motion norm and rate-ratio bounds. This gives a $.21$\,m/s
minimax root-risk lower bound with uniform rows.

\paragraph{Two-node special case.}
Using $(r_i,q_i)\in\{-.025,.175\}^2$ at each of the same
depths gives the fixture perturbed in Appendix Y. Its views
span $10$\,ms (\texttt{check\_antipodal\_alias.py}).

\paragraph{Finite-noise bound for random-direction motion.}
A separate pair gives root risk above $.389$\,m/s at $.1$\,px.
Use Appendix W's lateral coordinates, depths $4.4,4.7,5.0$,
nominal mounts, zero clocks, $\bar a=.1e_3$, $\omega=.005e_3$,
$B=0$ and $\bar v_\pm=\pm.02e_3$. Its spread is $.11$\,m.
The motion norm is $.1$ and the rate ratio $.05$. The world
lies in the support closure with strict depth and row margins.

At a common time the projection numerators $N_j(t)$ agree;
$\|(x,y)\|<.2$\,m and angular rate at most $.01$ give
$|N_j'|\leq.002$. The denominators are $D_\pm=z\mp.02t-.05t^2$.
On the raster $D_\pm\geq4.3775$ and $|\partial_t\pi_j|<.0015$,
with unique roots $|t|<.026$.

For each point and view, start at $s=.026$ and tighten
the time bound twice by replacing $s$ with
$(|N_1(0)|+.002s)/(z-.02s-.05s^2)$.
For the resulting bound let $h=z-.02s-.05s^2$ and
$B_j=|N_j(0)|+.002s$. On this interval,
\[
 A_j=\frac{.04s B_j}{h^2},\qquad
 M_j=\frac{.002}{h}+\frac{B_j(.02+.1s)}{h^2}
\]
bound the shared-time projection gap and within-state
time derivative, respectively. In particular the scan-root
gap is at most $A_1/(1-M_1)$, and the second coordinate
gap at most $A_2+M_2 A_1/(1-M_1)$.
Summing the squared bounds for all 96 coordinates and
multiplying by $(768/.1)^2$ gives the squared
noise-normalized mean distance $d^2<.451^2$ for each layout.
The rational calculation is in
\texttt{check\_generator\_risk.py}.

\paragraph{Gaussian squared-error loss.}
For the two states with equal prior, the sufficient scalar
observation satisfies $X\mid\pm\sim N(\pm d/2,1)$.
The posterior mean for the metric target is
$.4 e_3\tanh(dX/2)$. Thus the squared Bayes risk is at least
\[
 .16\bigl[1-d^2/4-d^4/16\bigr] > (.389)^2\,(\mathrm{m/s})^2,
\]
using $|\tanh x|\leq|x|$ and $\mathbb E X^2=1+d^2/4$.
The maximum risk dominates this Bayes risk.

\paragraph{Risk at $.1$ pixels on the Table 1 products.}
Fix the nominal points and mounts and set all clocks and remaining
motion coefficients to zero except $\bar v_\pm=\pm w e_3$,
where $w=.004\nu_r$ and $V_{\max,r}=20w$\,m/s.
These states belong to the same products. For an observation with
fixed projection numerators $n_1,n_2$ and nominal depth $z$,
the scan roots satisfy $t_\pm(z\mp wt_\pm)=n_1$.
On the raster $|t_\pm|\leq |n_1|/(z-w/2)$, and
$z(t_+-t_-)=w(t_+^2+t_-^2)$.
The second coordinate equals $(n_2/n_1)t_\pm$ for $n_1\ne0$.
When $n_1=0$, both coordinates agree. Thus the squared
normalized image difference per observation is bounded by
\[
 \frac{4w^2 n_1^2(n_1^2+n_2^2)}{z^2(z-w/2)^4}.
\]
Summing the bounds and scaling by $(768/.1)^2$ bounds
the squared Gaussian distance $d^2$. The source package's
rational checker certifies $d<.000027$ for Far and
$d<.02072$ for Near, on every layout.
The equal-prior Gaussian risk bound above gives
$R^*\geq V_{\max,r}^2(1-d^2/4-d^4/16)$, and zero
prediction gives $R^*\leq V_{\max,r}^2$.
The risk ratio exceeds $.9999999998$ for Far and $.99989$
for Near. Both endpoints have zero clock offsets, so these
bounds also apply when true clocks are supplied.

\nobalance
\section{Exposure-time spread and velocity}
\label{app:Y}
\renewcommand{\thetable}{Y.\arabic{table}}
\setcounter{table}{0}

\paragraph{Exact ambiguity at two exposure times.}
In the random-direction support, set $a=-.12$, $b=.195$,
$k=a+b$, $B=.1e_3$, $\omega=-kB/2$, and $u=.014$.
For $r,s\in\{a,b\}$ and $z\in\{4.35,4.65,4.95\}$\,m use
$R_*=R(a)=R(b)$ and
\[
\begin{aligned}
\bar v_\pm &= R_*^T u e_1\pm.05k e_3,\\
\bar a_\pm&=\mp.1e_3,\\
P_\pm&=R_*^T\bigl(r(z+u),.05-sz,z\bigr)^T\pm.05ab e_3 .
\end{aligned}
\]
At either node both rotated displacements equal
$(r(z+u)-ut,.05-sz,z)^T$. Anchors expose at $r$. Positive
and negative auxiliaries expose at $s$ and $k-s$, with
clocks zero and $k$, respectively. Each raster has unique
roots, depth above $4.33$\,m and $|\pi_1'|<.052$.
The image laws coincide even with supplied clocks.
The velocity gap is $.15$\,m/s, giving minimax root risk
at least $.075$\,m/s at any noise despite scene spread above
$.23$\,m and exposure span $15.75$\,ms. The source checker
verifies random-direction support and the branch.

All active anchor--auxiliary midpoints equal $k/2$.
Appendix P's upper bound therefore certifies zero
profiled velocity information. Its constant acceleration
step $-2w/k$ uses the same point shift $ab w/k$ for every
point in this two-node fixture.

\paragraph{Near ambiguity with distinct rows.}
For Appendix X's two-node states, index depths by
$d=0,1,2$ and put $j=2d+\mathbf 1\{r\ne s\}$,
$\delta=(2j-5)/1000$. Add $R_*^T(z\delta,-z\delta,0)^T$
to both point clouds. Each camera has all six shifts at
each original node $t_0$. At $t_0$ the scanned coordinate
shifts by $\pm\delta$. Since $|\pi_j'|<.04$ still holds,
the new root is within $|\delta|/24$ of $t_0\pm\delta$.
Row gaps exceed $1.2$\,px and exposure span exceeds
$10$\,ms. Each cluster spans at most $.53$\,ms.
The centred RMS spread remains above $.20$\,m.

At either root, with $g=|\delta|/.96$, the denominator
difference is at most $2mg(.2+g)$ and the numerator
derivative is below $.072$. Bound the shared-time projection
gap using these values and the local depth bound
$z-mg(.2+g)$, then apply Appendix X's contraction step.
The rational check \texttt{check\_distinct\_rows.py} gives
Gaussian distance below $.376$ at $.1$\,px and root risk
above $.392$\,m/s for each layout with shared clocks supplied.
Taking the shifts arbitrarily small while retaining distinct
rows gives exact minimax root risk $.4$\,m/s at any positive
noise when distinctness alone is required.

\paragraph{Random-direction clock comparison.}
In the prespecified confirmation, both true clocks are
supplied on 256 independent random-direction worlds.
Fits retain 60 unknowns, 96 observations per layout,
two starts and 300 evaluations per start. On outputs
projected onto the $.4$\,m/s speed ball, the simultaneous
$95\%$ upper endpoints on complementary mean excess
loss versus $++$, $--$ and zero are
$-.01945/-.02598/-.03054$\,(m/s)$^2$ by the L.12
e-value inversion (bound $.65$, allocation $.05/3$).

The matched free-clock comparison adds the two fitted
clock coordinates on the same inputs and keeps all
remaining fit settings. The projected free-clock upper
endpoints are $-.02850/-.02959/-.02167$\,(m/s)$^2$.

\begin{table}[H]
\centering\small
\caption{Matched random-direction RMSE (m/s).
The shared zero estimate gives $.285$\,m/s.}
\begin{tabular}{@{}llrrr@{}}
\toprule
Estimate & Clocks & $++$ & $--$ & $+-$\\
\midrule
Raw & Free & .375 & .374 & .226\\
Raw & True & .321 & .332 & .202\\
Projected & Free & .302 & .303 & .205\\
Projected & True & .267 & .279 & .184\\
\bottomrule
\end{tabular}
\end{table}

Supplying clocks reduces raw MSE for $++$, $--$ and
$+-$ by $.03762/.02961/.01043$\,(m/s)$^2$. Define
$G_R=L_{+-}-L_R$. The paired change in $G_R$
from supplying clocks is $+.02719/+.01918$ raw and
$+.01155/+.00610$ projected for $++/--$.
Nominal simultaneous $95\%$ intervals in $(\mathrm{m/s})^2$
are $[.01077,.04347]$ and $[.00559,.03320]$ raw,
and $[.00167,.02165]$ and $[-.00299,.01525]$ projected.
We use percentile intervals from 20,000 paired resamples
and Bonferroni correction across the four interactions.
All selected fits stop before the budget and have
projected gradient at most $10^{-5}$. Free-clock bound
hits occur in $65/60/1$ fits for $++/--/+-$.

\phantomsection\addcontentsline{toc}{section}{References}
\noindent Reference numbers match the submitted main paper. We list works cited in this supplementary material.